\documentclass{article}
\usepackage[final,nonatbib]{neurips_2018}
\usepackage{xspace}
\usepackage[square,numbers]{natbib}
\usepackage{listings}
\usepackage[utf8]{inputenc} 
\usepackage[T1]{fontenc}    
\usepackage[dvipsnames]{xcolor}
\usepackage[colorlinks,linkcolor=Aquamarine,urlcolor=violet,citecolor=YellowOrange,anchorcolor=violet]{hyperref}     
\usepackage{url}            
\usepackage{booktabs}       
\usepackage{amsfonts}       
\usepackage{nicefrac}       
\usepackage{microtype}      
\usepackage{multirow}
\usepackage{float}
\usepackage{graphicx}
\usepackage{dsfont}
\usepackage{bm}
\usepackage{xcolor}
\usepackage{wrapfig}
\usepackage{tikz}
\usepackage{amsmath,amsfonts,amssymb}
\usepackage{mathtools}
\usepackage{amsthm}
 \usepackage{subfig}
\usepackage{algorithmic}
\usepackage{epstopdf}
\usepackage[ruled,vlined]{algorithm2e}
\usepackage{mathrsfs}
\usepackage{rotating}
\usepackage[backgroundcolor = White,textwidth=\marginparwidth]{todonotes}
\def\half{{\textstyle\frac{1}{2}}}

\def\vec#1{\mathchoice%
	{\mbox{\boldmath $\displaystyle\bf#1$}}
	{\mbox{\boldmath $\textstyle\bf#1$}}
	{\mbox{\boldmath $\scriptstyle\bf#1$}}
	{\mbox{\boldmath $\scriptscriptstyle\bf#1$}}}
\def\v#1{\protect\vec #1}

\newcommand{\reals}{\mathbb R}
\newcommand{\D}{\mathcal D}
\newcommand{\A}{\mathcal A}
\newcommand{\B}{\mathcal B}

\newcommand{\C}{\mathcal C}

\newcommand{\G}{\mathcal G}

\newcommand{\ex}{\mathbb E}

\newcommand{\TODOK}[1]{
\ifmmode
\text{\textcolor{blue}{ }}
\else
\textcolor{blue}{ }
\fi
}
\newcommand{\TODOA}[1]{
\ifmmode
\text{\textcolor{red}{ }}
\else
\textcolor{red}{ }
\fi
}

\renewcommand{\vec}[1]{{\mathbf{#1}}}

\newtheorem{theorem}{Theorem}
\numberwithin{theorem}{section}

\makeatletter
\newcommand{\opnorm}{\@ifstar\@opnorms\@opnorm}
\newcommand{\@opnorms}[1]{%
  \left|\mkern-1.5mu\left|\mkern-1.5mu\left|
   #1
  \right|\mkern-1.5mu\right|\mkern-1.5mu\right|
}
\newcommand{\@opnorm}[2][]{%
  \mathopen{#1|\mkern-1.5mu#1|\mkern-1.5mu#1|}
  #2
  \mathclose{#1|\mkern-1.5mu#1|\mkern-1.5mu#1|}
}
\makeatother
\newcommand{\ALG}{FastGRNN\xspace}
\newcommand{\redSpace}{\vspace{-6mm}}

\newcommand{\algs}{FastRNN\xspace}
\newcommand{\alg}{FastGRNN\xspace}
\newcommand{\salg}{FastRNN\xspace}
\newcommand{\algfloat}{FastGRNN-LSQ\xspace}

\begin{document}

\makeatletter
\newcommand{\newreptheorem}[2]{\newtheorem*{rep@#1}{\rep@title}
\newenvironment{rep#1}[1]{\def\rep@title{#2 \ref*{##1}}\begin{rep@#1}}{\end{rep@#1}}
}
\makeatother

\newreptheorem{lemma}{Lemma}
\newreptheorem{theorem}{Theorem}
\newreptheorem{claim}{Claim}

\title{FastGRNN: A Fast, Accurate, Stable and Tiny Kilobyte Sized Gated Recurrent Neural Network}
\author{
Aditya Kusupati$^\dagger$, Manish Singh$^\mathsection$, Kush Bhatia$^\ddagger$, \\ \textbf{Ashish Kumar$^\ddagger$, Prateek Jain$^\dagger$ and Manik Varma$^\dagger$}\\
$^\dagger$Microsoft Research India\\
$^\mathsection$Indian Institute of Technology Delhi\\
$^\ddagger$University of California Berkeley\\
\texttt{\{t-vekusu,prajain,manik\}@microsoft.com}, \texttt{singhmanishiitd@gmail.com}\\ \texttt{kush@cs.berkeley.edu}, \texttt{ashish\_kumar@berkeley.edu}\\
}

\maketitle
\begin{abstract}

This paper develops the \salg and \alg algorithms to address the twin RNN limitations of inaccurate training and inefficient prediction. Previous approaches have improved accuracy at the expense of prediction costs making them infeasible for resource-constrained and real-time applications. Unitary RNNs have increased accuracy somewhat by restricting the range of the state transition matrix's singular values but have also increased the model size as they require a larger number of hidden units to make up for the loss in expressive power. Gated RNNs have obtained state-of-the-art accuracies by adding extra parameters thereby resulting in even larger models. \salg addresses these limitations by adding a residual connection that does not constrain the range of the singular values explicitly and has only two extra scalar parameters. \alg then extends the residual connection to a gate by reusing the RNN matrices to match state-of-the-art gated RNN accuracies but with a 2-4x smaller model. Enforcing \alg's matrices to be low-rank, sparse and quantized resulted in accurate models that could be up to 35x smaller than leading gated and unitary RNNs. This allowed \alg to accurately recognize the "Hey Cortana" wakeword with a 1 KB model and to be deployed on severely resource-constrained IoT microcontrollers too tiny to store other RNN models. \alg's code is available at~\citep{edgemlcode}.
\end{abstract}

\section{Introduction}
\label{sec:int}

\textbf{Objective}: This paper develops the \alg (an acronym for a Fast, Accurate, Stable and Tiny Gated Recurrent Neural Network) algorithm to address the twin RNN limitations of inaccurate training and inefficient prediction. \alg almost matches the accuracies and training times of state-of-the-art unitary and gated RNNs but has significantly lower prediction costs with models ranging from 1 to 6 Kilobytes for real-world applications.

\textbf{RNN training and prediction}: It is well recognized that RNN training is inaccurate and unstable as non-unitary hidden state transition matrices could lead to exploding and vanishing gradients for long input sequences and time series. An equally important concern for resource-constrained and real-time applications is the RNN's model size and prediction time. Squeezing the RNN model and code into a few Kilobytes could allow RNNs to be deployed on billions of Internet of Things (IoT) endpoints having just 2 KB RAM and 32 KB flash memory~\citep{Gupta17, Kumar17}. Similarly, squeezing the RNN model and code into a few Kilobytes of the 32 KB L1 cache of a Raspberry Pi or smartphone, could significantly reduce the prediction time and energy consumption and make RNNs feasible for real-time applications such as wake word detection~\citep{kepuska2009novel, chen2014small, chen2015query, sainath2015convolutional, siri}, predictive maintenance~\citep{susto2015machine, ahmad2017unsupervised}, human activity recognition~\citep{anguita2012, altun2010comparative}, {\it etc}.

\textbf{Unitary and gated RNNs}: A number of techniques have been proposed to stabilize RNN training based on improved optimization algorithms~\citep{pascanu2013, kanai2017}, unitary RNNs~\citep{arjovsky2016, jing2016, mhammedi2016, vorontsov2017, wisdom2016, zhang2018, jose2017} and gated RNNs~\citep{hochreiter1997, cho2014, collins2016}. While such approaches have increased the RNN prediction accuracy they have also significantly increased the model size. Unitary RNNs have avoided gradients exploding and vanishing by limiting the range of the singular values of the hidden state transition matrix. This has led to only limited gains in prediction accuracy as the optimal transition matrix might often not be close to unitary. Unitary RNNs have compensated by learning higher dimensional representations but, unfortunately, this has led to larger model sizes. Gated RNNs~\citep{hochreiter1997, cho2014, collins2016} have stabilized training by adding extra parameters leading to state-of-the-art prediction accuracies but with models that might sometimes be even larger than unitary RNNs.

\textbf{\salg}: This paper demonstrates that standard RNN training could be stabilized with the addition of a residual connection~\citep{he2016, srivastava2015, jaeger2007, bengio2013} having just 2 additional scalar parameters. Residual connections for RNNs have been proposed in~\citep{jaeger2007} and further studied in~\citep{bengio2013}. This paper proposes the \salg architecture and establishes that a simple variant of~\citep{jaeger2007, bengio2013} with learnt weighted residual connections~\eqref{eq:fastrnn} can lead to provably stable training and near state-of-the-art prediction accuracies with lower prediction costs than all unitary and gated RNNs. In particular, \salg's prediction accuracies could be: (a) up to 19\% higher than a standard RNN; (b) could often surpass the accuracies of all unitary RNNs and (c) could be just shy of the accuracies of leading gated RNNs. \salg's empirical performance could be understood on the basis of theorems proving that for an input sequence with $T$ steps and appropriate setting of residual connection weights: (a) \salg converges to a stationary point within $O(\nicefrac{1}{\epsilon^2})$ SGD iterations (see Theorem~\ref{thm:conv}), independent of $T$, while the {\em same analysis} for a standard RNN reveals an upper bound of $O(2^T)$ iterations and (b) \salg's generalization error bound is independent of $T$ whereas the {\em same proof technique} reveals an exponential bound for standard RNNs.

\textbf{\alg}: Inspired by this analysis, this paper develops the novel \alg architecture by converting the residual connection to a gate while reusing the RNN matrices. This allowed \alg to match, and sometimes exceed, state-of-the-art prediction accuracies of LSTM, GRU, UGRNN and other leading gated RNN techniques while having 2-4x fewer parameters. Enforcing \alg's matrices to be low-rank, sparse and quantized led to a minor decrease in the prediction accuracy but resulted in models that could be up to 35x smaller and fit in 1-6 Kilobytes for many applications. For instance, using a 1 KB model, \alg could match the prediction accuracies of all other RNNs at the task of recognizing the "Hey Cortana" wakeword. This allowed \alg to be deployed on IoT endpoints, such as the Arduino Uno, which were too small to hold other RNN models. On slightly larger endpoints, such as the Arduino MKR1000 or Due, \alg was found to be 18-42x faster at making predictions than other leading RNN methods.

\textbf{Contributions}: This paper makes two contributions. First, it rigorously studies the residual connection based \salg architecture which could often outperform unitary RNNs in terms of training time, prediction accuracy and prediction cost. Second, inspired by \salg, it develops the \alg architecture which could almost match state-of-the-art accuracies and training times but with prediction costs that could be lower by an order of magnitude. \salg and \alg's code can be downloaded from~\citep{edgemlcode}.

\section{Related Work}
\label{sec:rw}

\textbf{Residual connections}: Residual connections have been studied extensively in CNNs~\citep{he2016, srivastava2015} as well as RNNs~\citep{jaeger2007, bengio2013}. The Leaky Integration Unit architecture~\citep{jaeger2007} proposed residual connections for RNNs but were unable to learn the state transition matrix due to the problem of exploding and vanishing gradients. They therefore sampled the state transition matrix from a hand-crafted distribution with spectral radius less than one. This limitation was addressed in~\citep{bengio2013} where the state transition matrix was learnt but the residual connections were applied to only a few hidden units and with randomly sampled weights. Unfortunately, the distribution from which the weights were sampled could lead to an ill-conditioned optimization problem. In contrast, the \salg architecture leads to provably stable training with just two learnt weights connected to all the hidden units.

\textbf{Unitary RNNs}: Unitary RNNs~\citep{arjovsky2016, wisdom2016, mhammedi2016, jing2016, vorontsov2017, jose2017} stabilize RNN training by learning only well-conditioned state transition matrices. This limits their expressive power and prediction accuracy while increasing training time. For instance, SpectralRNN~\citep{zhang2018} learns a transition matrix with singular values in $1\pm \epsilon$. Unfortunately, the training algorithm converged only for small $\epsilon$ thereby limiting accuracy on most datasets. Increasing the number of hidden units was found to increase accuracy somewhat but at the cost of increased training time, prediction time and model size.

\textbf{Gated RNNs}: Gated architectures~\citep{hochreiter1997, cho2014, collins2016, jing2017} achieve state-of-the-art classification accuracies by adding extra parameters but also increase model size and prediction time. This has resulted in a trend to reduce the number of gates and parameters with UGRNN~\citep{collins2016} simplifying GRU~\citep{cho2014} which in turn simplifies LSTM~\citep{hochreiter1997}. \alg can be seen as a natural simplification of UGRNN where the RNN matrices are reused within the gate and are made low-rank, sparse and quantized so as to compress the model.

\textbf{Efficient training and prediction}: Efficient prediction algorithms have often been obtained by making sparsity and low-rank assumptions. Most unitary methods effectively utilize a low-rank representation of the state transition matrix to control prediction and training complexity~\citep{jing2016, zhang2018}. Sparsity, low-rank, and quantization were shown to be effective in RNNs~\citep{ye2017learning, narang2017, wang2017accelerating}, CNNs~\citep{han16}, trees~\citep{Kumar17} and nearest neighbour classifiers~\citep{Gupta17}. \alg\ builds on these ideas to utilize low-rank, sparse and quantized representations for learning kilobyte sized classifiers without compromising on classification accuracy. Other approaches to speed up RNN training and prediction are based on replacing sequential hidden state transitions by parallelizable convolutions~\citep{bradbury2016quasi} or on learning skip connections~\citep{campos2018} so as to avoid evaluating all the hidden states. Such techniques are complementary to the ones proposed in this paper and can be used to further improve \alg's performance.

\section{\salg and  \alg}
\label{sec:method}
\begin{figure}[t]
    \centering
    \subfloat[\salg\ - Residual Connection]{{\includegraphics[width=65mm]{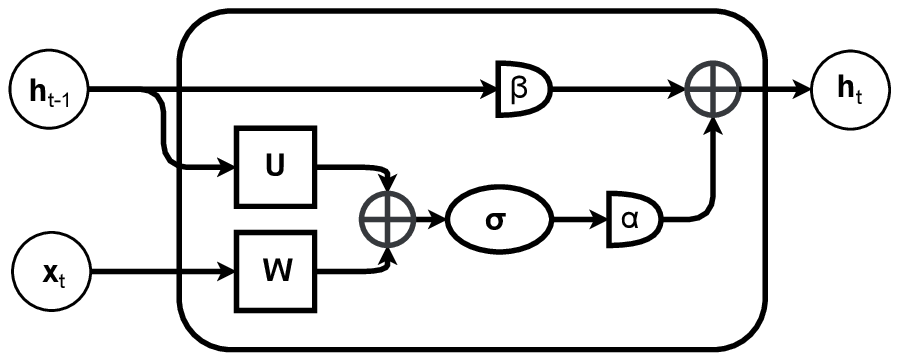}}}%
    \qquad
    \subfloat[\alg\ - Gate]{{\includegraphics[width=65mm]{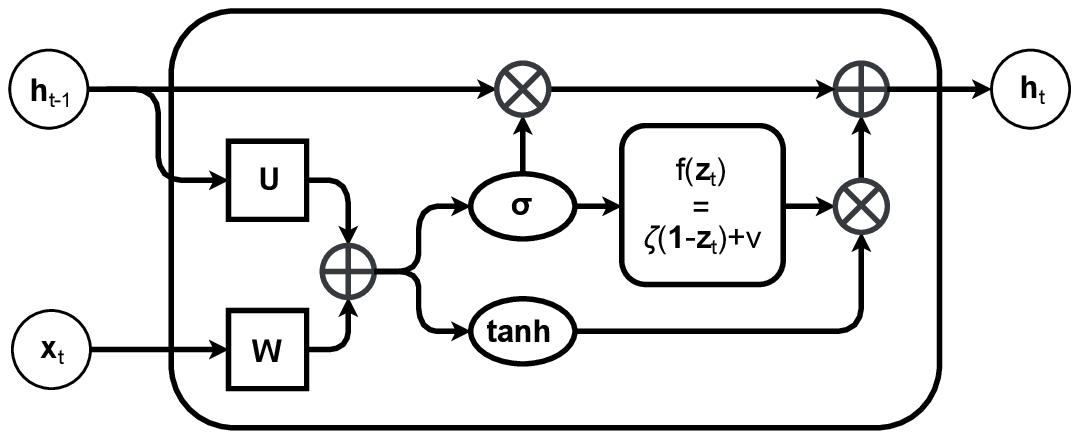}}}%
    \caption{Block diagrams for \salg (a) and \alg (b). \alg uses shared matrices $\v W$, $\v U$ to compute both the hidden state $\v h_t$ as well as the gate $\v z_t$.}%
    \label{fig:Archi}%
    \redSpace
\end{figure}

\textbf{Notation}: Throughout the paper, parameters of an RNN are denoted by matrices $\vec{W} \in \mathbb{R}^{\hat{D}\times D}, \vec{U} \in \mathbb{R}^{\hat{D}\times \hat{D}}$ and bias vectors $\v b \in \mathbb{R}^{\hat{D}}$, often using subscripts if multiple vectors are required to specify the architecture.  $\v a\odot \v b$ denotes the Hadamard product between $\v a$ and $\v b$, i.e., $(\v a\odot \v b)_i=\v a_i, \v b_i$. $\|\cdot \|_0$ denotes  the number of non-zeros entries in a matrix or vector. $\|\cdot \|_F, \|\cdot \|_2$ denotes the Frobenius and spectral norm of a matrix, respectively. Unless specified, $\|\cdot \|$ denotes $\|\cdot \|_2$ of a matrix or vector. $\v a^\top \v b=\sum_i a_i b_i$ denotes the inner product of $\v a$ and $\v b$.

Standard RNN architecture~\citep{rumelhart1986} is known to be {\em unstable} for training due to exploding or vanishing gradients and hence is shunned for more expensive gated architectures.

This paper studies the \salg architecture that is inspired by weighted residual connections~\citep{jaeger2007, he2016}, and shows that \salg can be significantly more stable and accurate than the standard RNN while preserving its prediction complexity. In particular, Section~\ref{sec:analysis}  demonstrates parameter settings for \salg that guarantee well-conditioned gradients as well as faster convergence rate and smaller generalization error than the standard RNN. This paper further strengthens \salg to develop the \alg architecture that is more accurate than unitary methods~\citep{arjovsky2016, zhang2018} and provides comparable accuracy to the state-of-the-art gated RNNs at 35x less computational cost (see Table~\ref{tab:results1}).

\subsection{\algs}

Let $\v X=[\v x_1, \dots, \v x_T]$ be the input data where $\v x_t\in \mathbb{R}^D$ denotes the $t$-th step feature vector. Then, the goal of multi-class RNNs is to learn a function $F:\mathbb{R}^{D\times T}\rightarrow \{1, \ldots, L\}$ that predicts one of $L$ classes for the given data point $\v X$.
Standard RNN architecture has a provision to produce an output at every time step, but we focus on the setting where each data point is associated with a single label that is predicted at the end of the time horizon $T$. Standard RNN maintains a vector of hidden state $\v h_t\in \mathbb{R}^{\hat{D}}$ which captures temporal dynamics in the input data, i.e.,
\begin{equation}
\v h_{t} = \tanh(\vec{W} \v x_t + \vec{U} \v h_{t-1} + \v b).
\end{equation}

As explained in the next section, learning $\vec{U}, \vec{W}$ in the above architecture is difficult as the gradient can have exponentially large (in $T$) condition number. Unitary methods explicitly control the condition number of the gradient but their training time can be significantly larger or the generated model can be less accurate.

Instead, \salg uses a simple weighted residual connection to stabilize the training by generating well-conditioned gradients. In particular, \salg updates the hidden state $\v h_t$ as follows:
\begin{align}
\vec{\tilde{h}}_{t}&= \sigma(\vec{W} \v x_t+\vec{U}\v h_{t-1}+\v b), \nonumber\\
\v h_{t} &= \alpha \vec{\tilde{h}}_{t}+\beta \v h_{t-1},\label{eq:fastrnn}
\end{align}
where $0\leq \alpha, \beta\leq1$ are trainable weights that are parameterized by the sigmoid function. $\sigma:\mathbb{R}\rightarrow \mathbb{R}$ is a non-linear function such as $\tanh$, sigmoid, or ReLU, and can vary across datasets. Given $\v h_T$, the label for a given point $\v X$ is predicted by applying a standard classifier, e.g., logistic regression to $\v h_T$.

Typically, $\alpha\ll 1$ and $\beta\approx 1-\alpha$, especially for problems with larger $T$. \salg updates hidden state in a controlled manner with  $\alpha, \beta$ limiting the extent to which the current feature vector $\v x_t$ updates the hidden state. Also, \salg has only $2$ more parameters than RNN and require only $\hat{D}$ more computations, which is a tiny fraction of per-step computation complexity of RNN. Unlike unitary methods~\citep{arjovsky2016, jing2017, zhang2018}, \salg does not introduce expensive structural constraints on $\v U$ and hence scales well to large datasets with standard optimization techniques~\citep{kingma2014}.

\subsubsection{Analysis}\label{sec:analysis}
This section shows how \salg addresses the issue of ill-conditioned gradients, leading to stable training and smaller  generalization error.  For simplicity, assume that the label decision function is one dimensional and is given by $f(\v X)=\v v^\top\v h_T$. Let $L(\v X,y;\v \theta)=L(f(\v X), y; \v \theta)$ be the logistic loss function for the given labeled data point $(\v X,y)$ and with parameters $\vec{\theta}=(\v W, \v U,\v v)$. Then, the gradient of $L$ w.r.t. $\v W, \v U, \v v$ is given by:
\begin{small}
\begin{align}
\frac{\partial L}{\partial \v U} &= \alpha \sum_{t=0}^{T} \v D_t \left(\prod_{k = t}^{T-1}(\alpha \v U^\top \v D_{k+1} + \beta \v I) \right)(\nabla_{\v h_T}L)\v h_{t-1}^\top,\\
\frac{\partial L}{\partial \v W} &= \alpha \sum_{t=0}^{T} \v D_t \left(\prod_{k = t}^{T-1}(\alpha \v U^\top \v D_{k+1} + \beta \v I) \right)(\nabla_{\v h_T}L)\v x_{t}^\top,
\frac{\partial L}{\partial \v v} = \frac{-y\exp{(-y\cdot\v v ^\top \v h_T)}}{1+\exp{(-y\cdot\v v ^\top \v h_T)}} \v h_T,
\end{align}
\end{small}where $\nabla_{\v h_T}L = -c(\v \theta)\cdot y \cdot \v v$, and $c(\v \theta) = \frac{1}{1+\exp{(y\cdot\v v^\top \v h_T)}}$. A critical term in the above expression is: $M(\v U)=\prod_{k = t}^{T-1}(\alpha \v U^\top \v D_{k+1} + \beta \v I)$, whose condition number, $\kappa_{M(\v U)}$, is bounded by:
\begin{align}
\kappa_{M(\v U)}\leq \frac{(1+\frac{\alpha}{\beta}\max_k \|\v U^\top \v D_{k+1}\|)^{T-t}}{(1-\frac{\alpha}{\beta}\max_k \|\v U^\top \v D_{k+1}\|)^{T-t}},\label{eq:kappa}
\end{align}where $\vec{D}_{k} = \text{diag}(\sigma'(\vec{W}\vec{x}_k+\vec{U}\vec{h}_{k-1}+\vec{b}))$ is the Jacobian matrix of the pointwise nonlinearity. Also if $\alpha=1$ and $\beta=0$, which corresponds to standard RNN, the condition number of $M(\v U)$ can be as large as $(\max_k \frac{\|\v U^\top \v D_{k+1}\|}{\lambda_{min}(\v U^\top \v D_{k+1})})^{T-t}$ where $\lambda_{min}(\v A)$ denotes the minimum singular value of $\v A$. Hence, gradient's condition number for the standard RNN can be exponential in $T$. This implies that, relative to the average eigenvalue, the gradient can explode or vanish in certain directions, leading to unstable training.

In contrast to the standard RNN, if $\beta\approx 1$ and $\alpha\approx 0$, then the condition number, $\kappa_{M(\v U)}$, for \algs\ is bounded by a small term. For example, if $\beta=1-\alpha$ and $\alpha=\frac{1}{T \max_k \|\v U^\top\v D_{k+1}\|}$, then $\kappa_{M(\v U)}=O(1)$. Existing unitary methods are also motivated by similar observation. But they attempt to control the $\kappa_{M(\v U)}$ by restricting the condition number, $\kappa_{\v U}$, of $\v U$ which can still lead to ill-conditioned gradients as $\v U^\top \v D_{k+1}$ might still be very small in certain directions. By using residual connections, \salg is able to address this issue, and hence have faster training and more accurate model than the state-of-the-art unitary RNNs.

Finally, by using the above observations and a careful perturbation analysis, we can provide the following convergence and generalization error bounds for \salg:
\begin{theorem}[Convergence Bound] \label{thm:conv}
	Let $[(\v X_1, y_1), \dots, (\v X_n, y_n)]$ be the given labeled sequential training data. Let $L(\v \theta)=\frac{1}{n}\sum_i L(\v X_i, y_i; \v \theta)$ be the loss function with $\v \theta=(\v W, \v U, \v v)$ be the parameters of \algs\ architecture \eqref{eq:fastrnn} with $\beta = 1-\alpha$ and $\alpha$ such that,
\vspace{-0.4mm}
  \begin{equation*}
    \alpha \leq \min\left( \frac{1}{4T\cdot|\D\|\vec{U}\|_2 -1| }, \frac{1}{4T\cdot R_\vec{U}},  \frac{1}{T\cdot |\|{\vec{U}}\|_2-1|} \right),
  \end{equation*}
  \vspace{-0.75mm}
  where $\D = \sup_{\theta, k}\| \vec{D}_k^\theta\|_2$. Then, randomized stochastic gradient descent \cite{ghadimi2013}, a minor variation of SGD, when applied to the data for a maximum of $M$ iteration outputs a solution $\widehat{\v \theta}$ such that:
  \vspace{-0.4mm}
\begin{equation*} 
	\mathbb{E}[\| \nabla_{\v \theta}L(\widehat{\v \theta})\|_2^2 \| ]\leq \mathcal{B}_M := \frac{\mathcal{O}(\alpha T)L(\v \theta_0)}{M} + \left( \bar{D} + \frac{4R_{\v W} R_{\v U} R_{\v v}}{\bar{D}}\right)\frac{\mathcal{O}(\alpha T)}{\sqrt{M}}\leq \epsilon,
	\end{equation*}
	\vspace{-0.75mm}
	where $R_{\v X} = \max_{\v X} \|\v X\|_F\ $ for $\v X = \{\v U, \v W, \v v\}$, $L(\v \theta_0)$ is the loss of the initial classifier, and the step-size of the $k$-th SGD iteration is fixed as: $\gamma_k = \min \left\{ \frac{1}{\mathcal{O}(\alpha T)}, \frac{\bar{D}}{T \sqrt{M}}\right\}, k \in [M],\ \ \bar{D}\geq 0.$ Maximum number of iterations is bounded by $M=O(\frac{\alpha T}{\epsilon^2}\cdot poly(L(\theta_0), R_{\v W} R_{\v U} R_{\v v}, \bar{D}))$, $\epsilon\geq 0$. 
\end{theorem}

\begin{theorem}[Generalization Error Bound] \label{thm:gen}
	  \cite{bartlett2002}  Let $\mathcal{Y}, \hat{\mathcal{Y}} \subseteq [0,1]$ and let $\mathcal{F}_T$ denote the class of \algs\ with $\|\v U\|_F \leq R_{\v U}, \|\v W\|_F\leq R_{\v W}$. Let the final classifier be given by $\sigma(\v v^\top \v h_T)$, $\|\v v\|_2 \leq R_{\v v}$ . Let $L:\mathcal{Y}\times  \hat{\mathcal{Y}} \to [0, B]$ be any $1$-Lipschitz loss function. Let $D$ be any distribution on $\mathcal{X}\times \mathcal{Y}$ such that $\|\v x_{it}\|_2 \leq R_x$ a.s. Let $0\leq \delta \leq 1$. For all $\beta = 1-\alpha$ and $\alpha$ such that,
\vspace{-0.4mm}
    \begin{equation*}
\alpha \leq \min\left( \frac{1}{4T\cdot|\D\|\vec{U}\|_2 -1| }, \frac{1}{4T\cdot R_\vec{U}}, \frac{1}{T\cdot |\|{\vec{U}}\|_2-1|} \right),
\end{equation*}
\vspace{-0.75mm}
      where $\D = \sup_{\theta, k}\| \vec{D}_k^\theta\|_2$, we have that with probability at least $1-\delta$, all functions $f \in \v v \circ \mathcal{F}_T$ satisfy,
      \vspace{-0.4mm}
	  \begin{equation*}
	    \ex_D[L(f(\v X), y)] \leq \frac{1}{{n}}\sum_{i=1}^n L(f(\v X_i), y_i) + \mathcal{C}\frac{O(\alpha T)}{\sqrt{n}} + B\sqrt{\frac{\ln(\frac{1}{\delta})}{n}},
	  \end{equation*}
	  \vspace{-0.75mm}
	  where $\mathcal{C} = R_{\vec{W}} R_{\vec{U}} R_{\vec{x}} R_{\vec{v}}$ represents the boundedness of the parameter matrices and the data.
\end{theorem}

The convergence bound states that if $\alpha=O(1/T)$ then the algorithm converges to a stationary point in {\em constant time} with respect to $T$ and polynomial time with respect  to all the other problem parameters. Generalization bound states that for $\alpha=O(1/T)$, the generalization error of \algs\ is {\em independent} of $T$. In contrast, {\em similar proof technique} provide exponentially poor (in $T$) error bound and convergence rate for standard RNN. But, this is an upper bound, so potentially significantly better error bounds for RNN might exist; matching lower bound results for standard RNN is an interesting research direction. Also, $O(T^2)$ generalization error bound can be argued using VC-dimension style arguments~\citep{anthony2009neural}. But such bounds hold for specific settings like binary $y$, and are independent of problem hardness parameterized by the size of the weight matrices ($R_{\v W}, R_{\v U}$).

Finally, note that the above analysis fixes $\alpha=O(1/T)$, $\beta=1-\alpha$, but in practice \salg learns $\alpha, \beta$ (which is similar to performing cross-validation on $\alpha, \beta$). However, interestingly, across datasets the learnt $\alpha, \beta$ values indeed display a similar scaling wrt $T$ for large $T$ (see Figure~\ref{fig:plt_ab}).

\subsection{\ALG}
While \algs controls the condition number of gradient reasonably well, its expressive power might be limited for some datasets. This concern is addressed by a novel architecture, \ALG, that uses a scalar weighted residual connection for each and every coordinate of the hidden state $\v h_t$. That is,
\begin{align}
\v z_t&= \sigma(\vec{W} \v x_t+\vec{U}\v h_{t-1}+\v b_z), \nonumber\\
\vec{\tilde{h}}_{t}&= \tanh(\vec{W} \v x_t+\vec{U}\v h_{t-1}+\v b_h), \nonumber\\
\v h_t&=  \left(\zeta(\v 1-\v z_t) + \nu \right)\odot\vec{\tilde{h}}_{t}+\v z_t\odot \v h_{t-1},\label{eq:fastgrnn}
\end{align}
where $0\leq \zeta, \nu\leq 1$ are trainable parameters that are parameterized by the sigmoid function, and $\sigma:\mathbb{R}\rightarrow \mathbb{R}$ is a non-linear function such as $\tanh$, sigmoid and can vary across datasets. Note that each coordinate of $\v z_t$ is similar to parameter $\beta$ in \eqref{eq:fastrnn} and  $\zeta (\v 1-\v z_t)+\nu$'s coordinates simulate $\alpha$ parameter; also if $\nu\approx0, \zeta\approx 1$ then it satisfies the intuition that $\alpha+\beta=1$. It was observed that across all datasets, this gating mechanism outperformed the simple vector extension of \salg where each coordinate of $\vec{\alpha}$ and $\vec{\beta}$ is learnt (see Appendix~\ref{sec:vectorfastrnn}).

\ALG computes each coordinate of gate $\v z_t$ using a non-linear function of $\v x_t$ and $\v h_{t-1}$. To minimize the number of parameters, \alg reuses the matrices $\v W$, $\v U$ for the vector-valued gating function as well. Hence, \alg's inference complexity is almost same as that of the standard RNN but its accuracy and training stability is on par with expensive gated architectures like GRU and LSTM.

{\bf Sparse low-rank representation}: \ALG further compresses the model size by using a low-rank and a sparse representation of the parameter matrices $\v W$, $\v U$. That is,
\begin{equation}
\label{eq:lr}
\vec{W} = \vec{W}^1 (\vec{W}^2)^\top,\ \vec{U} = \vec{U}^1 (\vec{U}^2)^\top,\ \|\vec{W}^i\|_0 \leq s_w^i,\ \|\vec{U}^i\|_0 \leq s_u^i,\ i=\{1,2\},
\end{equation}
where $\vec{W}^1 \in \reals^{\hat{D} \times r_w}, \vec{W}^2 \in \reals^{D \times r_w}$, and $\vec{U}^1, \vec{U}^2 \in \reals^{\hat{D} \times r_u}$.
Hyperparameters $r_w, s_w, r_u, s_u$ provide an efficient way to control the \emph{accuracy-memory} trade-off for \ALG and are typically set via fine-grained validation. In particular, such compression is critical for \alg model to fit on resource-constrained devices. Second, this low-rank representation brings down the prediction time by reducing the cost at each time step from $\mathcal{O}(\hat{D}(D+\hat{D}))$ to $\mathcal{O}(r_w(D+\hat{D}) + r_u\hat{D})$. This enables \ALG\ to provide on-device prediction in real-time on battery constrained devices.

\subsubsection{Training \ALG}
The parameters for \ALG: $\v \Theta_{\text{\ALG}} = (\vec{W}^i, \vec{U}^i, \v b_h, \v b_z, \zeta, \nu)$ are trained jointly using projected batch stochastic gradient descent (b-SGD) (or other stochastic optimization methods) with typical batch sizes ranging from $64-128$. In particular, the optimization problem is given by:{\small
\begin{align}
\min_{\Theta_{\text{\ALG}}, \|\vec{W}^i\|_0 \leq s_w^i, \|\vec{U}^i\|_0 \leq s_u^i, i\in \{1,2\}}\mathcal{J}(\Theta_{\text{\ALG}}) = \frac{1}{n}\sum_j L(\v X_j, y_j; \Theta_{\text{\ALG}})
\end{align}}where $L$ denotes the appropriate loss function (typically softmax cross-entropy). The training procedure for \ALG\ is divided into 3 stages:

\textbf{(I) Learning low-rank representation (L)}: In the first stage of the training, \ALG\ is trained for $e_1$ epochs with the model as specified by \eqref{eq:lr} using b-SGD. This stage of optimization ignores the sparsity constraints on the parameters and learns a low-rank representation of the parameters.

\textbf{(II) Learning sparsity structure (S)}: \ALG\ is next trained for $e_2$ epochs using b-SGD, projecting the parameters onto the space of sparse low-rank matrices after every few batches while maintaining support between two consecutive projection steps.
This stage, using b-SGD with Iterative Hard Thresholding (IHT), helps \ALG\ identify the correct support for parameters $(\vec{W}^i,\vec{U}^i)$.

\textbf{(III) Optimizing with fixed parameter support}: In the last stage, \ALG\ is trained for $e_3$ epochs with b-SGD while freezing the support set of the parameters.

In practice, it is observed that $e_1=e_2=e_3=100$ generally leads to the convergence of \ALG\ to a good solution. Early stopping is often deployed in stages (II) and (III) to obtain the best models.

\subsection{Byte Quantization (Q)}
\alg further compresses the model by quantizing each element of $\v W^i, \v U^i$, restricting them to at most one byte along with byte indexing for sparse models. However, simple integer quantization of $\v W^i, \v U^i$ leads to a large loss in accuracy due to gross approximation. Moreover, while such a quantization reduces the model size, the prediction time can still be large as non-linearities will require all the hidden states to be floating point. \alg overcomes these shortcomings by training $\v W^i$ and $\v U^i$ using {\em piecewise-linear} approximation of the non-linear functions, thereby ensuring that all the computations can be performed with integer arithmetic. During training, \alg replaces the non-linear function in \eqref{eq:fastgrnn} with their respective approximations and uses the above mentioned training procedure to obtain $\v \Theta_{\text{\ALG}}$. The floating point parameters are then jointly quantized to ensure that all the relevant entities are integer-valued and the entire inference computation can be executed efficiently with integer arithmetic without a significant drop in accuracy. For instance, Tables~\ref{tab:mkr1000},~\ref{tab:due} show that on several datasets \alg models are 3-4x faster than their corresponding \alg-Q models on common IoT boards with no floating point unit (FPU). \algfloat, \alg "minus" the Low-rank, Sparse and Quantized components, is the base model with no compression.

\section{Experiments}
\label{sec:exp}
\begin{table}[b]
\centering
\redSpace
\makebox[0pt][c]{\parbox{1\textwidth}{%
    \begin{minipage}[c]{0.5\hsize}\centering
    \captionsetup{font=small}
     \caption{Dataset Statistics}

  \resizebox{0.98\linewidth}{!}
  {
       \begin{tabular}{lcccc}
    \toprule
    Dataset & \#Train &\#Features &\multicolumn{1}{c}{\begin{tabular}[c]{@{}c@{}}\#Time\\ Steps\end{tabular}}& \#Test  \\
    \midrule
    Google-12 &   \phantom{0}22,246 &  \phantom{0}3,168&\phantom{0}99&  \phantom{00}3,081 \\ 
    Google-30 &   \phantom{0}51,088 &  \phantom{0}3,168&\phantom{0}99&  \phantom{00}6,835  \\ 
    Wakeword-2 &    195,800 &  \phantom{0}5,184&162&  \phantom{0}83,915  \\ 
    Yelp-5 &    500,000 & 38,400&300& 500,000   \\
    HAR-2 &   \phantom{00}7,352&   \phantom{0}1,152 &128& \phantom{00}2,947  \\ 
    Pixel-MNIST-10 &  \phantom{0}60,000&   \phantom{00}784 &784&  \phantom{0}10,000  \\ 
    PTB-10000 & 929,589 &---&300&  \phantom{0}82,430 \\
     DSA-19 &  \phantom{00}4,560 &  \phantom{0}5,625&125&  \phantom{00}4,560  \\ 

    \bottomrule
  \end{tabular}
  \label{tab:stat}
  }
    \end{minipage}
    \hfill
    \begin{minipage}[b]{0.5\hsize}\centering
    \captionsetup{font=small}
       \caption{PTB Language Modeling - 1 Layer}

  \resizebox{0.98\linewidth}{!}
  {
  \begin{tabular}{@{}lcccc@{}}
\toprule
Method                   & \multicolumn{1}{c}{\begin{tabular}[c]{@{}c@{}}Test\\ Perplexity\end{tabular}} & \multicolumn{1}{c}{\begin{tabular}[c]{@{}c@{}}Train\\ Perplexity\end{tabular}} & \multicolumn{1}{c}{\begin{tabular}[c]{@{}c@{}}Model\\ Size (KB)\end{tabular}} & \multicolumn{1}{c}{\begin{tabular}[c]{@{}c@{}}Train\\ Time (min)\end{tabular}} \\ \midrule
RNN                      & 144.71                                                                       & \phantom{0}68.11                                                                          & \phantom{0}129                                                                           & \textbf{\phantom{0}9.11}                                                                   \\ 
\salg     & 127.76\textsuperscript{+}                                                                        & 109.07                                                                         & \phantom{0}513                                                                           & 11.20                                                                           \\
\algfloat & \textbf{115.92}                                                                        & \phantom{0}89.58                                                                          & \phantom{0}513                                                                           & 12.53                                                                           \\
\alg      & 116.11                                                                        & \phantom{0}81.31                                                                          & \textbf{\phantom{00}39}                                                                   & 13.75                                                                          \\ 
SpectralRNN                   & 130.20                                                                       & \phantom{0}65.42                                                                          & \phantom{0}242                                                                           & ---                                                                            \\ 
UGRNN                    & 119.71                                                                        & \phantom{0}65.25                                                                          & \phantom{0}256                                                                           & 11.12                                                                          \\
LSTM                     & 117.41                                                                        & \phantom{0}69.44                                                                          & 2052                                                                          & 13.52                                                                           \\ \bottomrule
\end{tabular}
  \label{tab:ptb}
  }
    \end{minipage}

}}
\end{table}

\textbf{Datasets}: \salg and \alg's performance was benchmarked on the following IoT tasks where having low model sizes and prediction times was critical to the success of the application: (a) Wakeword-2~\citep{wakeword} - detecting utterances of the "Hey Cortana" wakeword; (b) Google-30~\citep{warden2017} and Google-12 - detection of utterances of 30 and 10 commands plus background noise and silence and (c) HAR-2~\citep{anguita2012} and DSA-19~\citep{altun2010comparative} - Human Activity Recognition (HAR) from an accelerometer and gyroscope on a Samsung Galaxy S3 smartphone and Daily and Sports Activity (DSA) detection from a resource-constrained IoT wearable device with 5 Xsens MTx sensors having accelerometers, gyroscopes and magnetometers on the torso and four limbs. Traditional RNN tasks typically do not have prediction constraints and are therefore not the focus of this paper. Nevertheless, for the sake of completeness, experiments were also carried out on benchmark RNN tasks such as language modeling on the Penn Treebank (PTB) dataset~\citep{marcus1993building}, star rating prediction on a scale of 1 to 5 of Yelp reviews~\citep{yelp} and classification of MNIST images on a pixel-by-pixel sequence~\citep{lecun1998, le2015simple}.

All datasets, apart from Wakeword-2, are publicly available and their pre-processing and feature extraction details are provided in Appendix~\ref{sec:datasetinfo}. The publicly provided training set for each dataset was subdivided into $80\%$ for training and $20\%$ for validation. Once the hyperparameters had been fixed, the algorithms were trained on the full training set and results were reported on the publicly available test set. Table~\ref{tab:stat} lists the statistics of all datasets.

\begin{table}[b]
\redSpace
\centering
\makebox[0pt][c]{\parbox{1\textwidth}{%
    \begin{minipage}[c]{1\hsize}\centering
    \captionsetup{font=small}
     \caption{ \alg had up to 35x smaller models than leading RNNs with almost no loss in accuracy}

  \resizebox{0.98\linewidth}{!}
  {
\scalebox{0.73}{
\begin{tabular}{@{}r|lcccccccccc@{}}
\toprule
                          & Dataset                  & \multicolumn{3}{c}{Google-12}                                                                                                                                                                                              & \multicolumn{3}{c}{Google-30}                                                                                                                                                                                              & \multicolumn{3}{c}{Wakeword-2}                                                                                                                                                                                           \\ \midrule
                          & Method                   & \begin{tabular}[c]{@{}c@{}}Accuracy\\ (\%)\end{tabular} & \multicolumn{1}{c}{\begin{tabular}[c]{@{}c@{}}Model \\ Size (KB)\end{tabular}} & \multicolumn{1}{c}{\begin{tabular}[c]{@{}c@{}}Train \\ Time (hr)\end{tabular}} & \begin{tabular}[c]{@{}c@{}}Accuracy\\ (\%)\end{tabular} & \multicolumn{1}{c}{\begin{tabular}[c]{@{}c@{}}Model \\ Size (KB)\end{tabular}} & \multicolumn{1}{c}{\begin{tabular}[c]{@{}c@{}}Train \\ Time (hr)\end{tabular}} & \begin{tabular}[c]{@{}c@{}}F1 \\ Score\end{tabular} & \multicolumn{1}{c}{\begin{tabular}[c]{@{}c@{}}Model \\ Size (KB)\end{tabular}} & \multicolumn{1}{c}{\begin{tabular}[c]{@{}c@{}}Train \\ Time(hr)\end{tabular}} \\ \midrule
                          & RNN                      & 73.25                                                   & \phantom{00}56                                                                             & \phantom{00}1.11                                                       & 80.05                                                   & \phantom{00}63                                                                             & \phantom{0}2.13                                                       & 89.17                                               & \phantom{0}8                                                                              & \textbf{\phantom{0}0.28}                                                                 \\ \midrule
\multirow{3}{*}{\begin{sideways}\vspace{1.5mm}Proposed \end{sideways}} & \salg     & 92.21\textsuperscript{+}                                                   & \phantom{00}56                                                                             & \textbf{\phantom{00}0.61}                                              & 91.60\textsuperscript{+}                                                   & \phantom{00}96                                                                             & \textbf{\phantom{0}1.30}                                              & 97.09                                               & \phantom{0}8                                                                              & \phantom{0}0.69                                                                          \\
                          & \algfloat & \textbf{93.18}                                          & \phantom{00}57                                                                             & \phantom{00}0.63                                                      & \textbf{92.03}                                          & \phantom{00}45                                                                             & \phantom{0}1.41                                                       & \textbf{98.19}                                      & \phantom{0}8                                                                              & \phantom{0}0.83                                                                          \\
                          & \alg      & 92.10                                                   & \textbf{\phantom{0}5.5}                                                                   & \phantom{00}0.75                                                       & 90.78                                                   & \textbf{\phantom{0}6.25}                                                                  & \phantom{0}1.77                                                       & 97.83                                               & \textbf{\phantom{0}1}                                                                     & \phantom{0}1.08                                                                          \vspace{1mm}\\  \midrule
\multirow{4}{*}{\begin{sideways}Unitary\end{sideways}}  & SpectralRNN              & 91.59                                                   & \phantom{0}228                                                                            & \phantom{0}19.00                                                      & 88.73                                                   & \phantom{0}128                                                                            & 11.00                                                      & 96.75                                               & 17                                                                             & \phantom{0}7.00                                                                          \\
                          & EURNN                     & 76.79                                                   & \phantom{0}210                                                                            & 120.00                                                     & 56.35                                                   & \phantom{0}135                                                                            & 19.00                                                      & 92.22                                               & 24                                                                             & 69.00                                                                         \\
                          & oRNN                     & 88.18                                                   & \phantom{0}102                                                                            & \phantom{0}16.00                                                      & 86.95                                                   & \phantom{0}120                                                                            & 35.00                                                      & ---                                                 & ---                                                                            & ---                                                                           \\
                          & FactoredRNN              & 53.33                                                   & 1114                                                                           & \phantom{00}7.00                                                       & 40.57                                                   & 1150                                                                           & \phantom{0}8.52                                                       & ---                                                 & ---                                                                            & ---                                                                           \\ \midrule
\multirow{3}{*}{\begin{sideways}Gated\end{sideways}}    & UGRNN                    & 92.63                                                   & \phantom{00}75                                                                             & \phantom{0}0.78                                                       & 90.54                                                   & \phantom{0}260                                                                            & \phantom{0}2.11                                                       & 98.17                                               & 16                                                                             & \phantom{0}1.00                                                                          \\
                          & GRU                      & 93.15                                                   & \phantom{0}248                                                                            & \phantom{0}1.23                                                       & 91.41                                                   & \phantom{0}257                                                                            & \phantom{0}2.70                                                       & 97.63                                               & 24                                                                             & \phantom{0}1.38                                                                          \\
                          & LSTM                     & 92.30                                                   & \phantom{0}212                                                                            & \phantom{0}1.36                                                       & 90.31                                                   & \phantom{0}219                                                                            & \phantom{0}2.63                                                       & 97.82                                               & 32                                                                             & \phantom{0}1.71                                                                          \\ \bottomrule
\end{tabular}}
  \label{tab:results1}
  }
    \end{minipage}
\hfill

\begin{minipage}[c]{1\hsize}\centering
\vspace{3mm}
  \resizebox{0.98\linewidth}{!}
  {
\scalebox{0.54}{
\begin{tabular}{@{}r|lccc|ccc|ccc|ccc@{}}
\toprule
                          & Dataset                  & \multicolumn{3}{c}{Yelp-5}                                                                                                                                                                                                 & \multicolumn{3}{c}{HAR-2} & \multicolumn{3}{c}{DSA-19}                                                                                                                                                                                              & \multicolumn{3}{c}{Pixel-MNIST-10}                                                                                                                                                                                         \\ \midrule
                          & Method                   & \begin{tabular}[c]{@{}c@{}}Accuracy\\ (\%)\end{tabular} & \multicolumn{1}{c}{\begin{tabular}[c]{@{}c@{}}RNN Model \\ Size (KB)\end{tabular}} & \multicolumn{1}{c}{\begin{tabular}[c]{@{}c@{}}Train \\ Time (hr)\end{tabular}} & \begin{tabular}[c]{@{}c@{}}Accuracy\\ (\%)\end{tabular} & \multicolumn{1}{c}{\begin{tabular}[c]{@{}c@{}}Model \\ Size (KB)\end{tabular}} & \multicolumn{1}{c}{\begin{tabular}[c]{@{}c@{}}Train \\ Time (hr)\end{tabular}} & \begin{tabular}[c]{@{}c@{}}Accuracy \\ (\%)\end{tabular} & \multicolumn{1}{c}{\begin{tabular}[c]{@{}c@{}}Model \\ Size (KB)\end{tabular}} & \multicolumn{1}{c}{\begin{tabular}[c]{@{}c@{}}Train \\ Time (min)\end{tabular}} & \begin{tabular}[c]{@{}c@{}}Accuracy \\ (\%)\end{tabular} & \multicolumn{1}{c}{\begin{tabular}[c]{@{}c@{}}Model \\ Size (KB)\end{tabular}} & \multicolumn{1}{c}{\begin{tabular}[c]{@{}c@{}}Train \\ Time (hr)\end{tabular}}\\ \midrule
                          & RNN                      & 47.59                                                   & 130                                                                            & {\textbf{\phantom{0}3.33}}                                              & 91.31                                                   & \phantom{0}29                                                                             & {0.11}  & 71.68                                                   & \phantom{00}20                                                                             & \textbf{1.11}                                                       & 94.10                                                    & \phantom{0}71                                                                             & \phantom{0}45.56                                                                          \\ \midrule
\multirow{3}{*}{\begin{sideways}Proposed\end{sideways}} & \salg     & 55.38                                                   & 130                                                                            & {\phantom{0}3.61}                                                       & 94.50\textsuperscript{+}                                                   & \phantom{0}29                                                                             & {\textbf{0.06}}  & 84.14                                                   & \phantom{00}97                                                                             & {1.92}                                             & 96.44                                                    & 166                                                                            & \phantom{0}15.10                                                                          \\
                          & \algfloat & \textbf{59.51}                                          & 130                                                                            & {\phantom{0}3.91}                                                       & 95.38                                                   & \phantom{0}29                                                                             & {0.08}    & \textbf{85.00}                                                   & \phantom{0}208                                                                             & {2.15}                                                    & \textbf{98.72}                                           & \phantom{0}71                                                                             & \textbf{\phantom{0}12.57}                                                                 \\
                          & \alg      & 59.43                                                   & \textbf{\phantom{00}8}                                                                     & {\phantom{0}4.62}                                                       & \textbf{95.59}                                          & \phantom{00}3                                                                              & {0.10}  & 83.73                                                   & \textbf{3.25}                                                                             & 2.10                                                     & 98.20                                                    & \textbf{\phantom{00}6}                                                                     & \phantom{0}16.97                                                                          \vspace{1mm}\\ \midrule
\multirow{4}{*}{\begin{sideways}Unitary\end{sideways}}  & SpectralRNN              & 56.56                                                   & \phantom{0}89                                                                             & {\phantom{0}4.92}                                                       & 95.48                                                   & 525                                                                            & {0.73} & 80.37                                                   & \phantom{00}50                                                                             & 2.25                                                       & 97.70                                                   & \phantom{0}25                                                                             & ---                                                                            \\
                          & EURNN                     & 59.01                                                   & 122                                                                            & {72.00}                                                      & 93.11                                                   & \phantom{0}12                                                                             & {0.84}   & ---                                                   & ---                                                                             & ---                                                     & 95.38                                                    & \phantom{0}64                                                                             & 122.00                                                                         \\
                          & oRNN                     & ---                                                     & ---                                                                            & {---}                                                        & 94.57                                                   & \phantom{0}22                                                                             & {2.72}  & 72.52                                                   & \phantom{00}18                                                                             & ---                                                      & 97.20                                                   & \phantom{0}49                                                                             & ---                                                                            \\
                          & FactoredRNN              & ---                                                     & ---                                                                            & {---}                                                        & 78.65                                                   & \textbf{\phantom{00}1}                                                                     & {0.11}  & 73.20                                                   & 1154                                                                             & ---                                                      & 94.60                                                   & 125                                                                            & ---                                                                            \\ \midrule
\multirow{3}{*}{\begin{sideways}Gated\end{sideways}}    & UGRNN                    & 58.67                                                   & 258                                                                            & {\phantom{0}4.34}                                                       & 94.53                                                   & \phantom{0}37                                                                             & {0.12}       & 84.74                                                   & \phantom{0}399                                                                             & {2.31}                                                 & 97.29                                                    & \phantom{0}84                                                                             & 15.17                                                                          \\
                          & GRU                      & 59.02                                                   & 388                                                                            & {\phantom{0}8.12}                                                       & 93.62                                                   & \phantom{0}71                                                                             & {0.13}       & 84.84                                                   & \phantom{0}270                                                                             & {2.33}                                                 & 98.70                                                    & 123                                                                            & 23.67                                                                          \\
                          & LSTM                     & 59.49                                                   & 516                                                                            & {\phantom{0}8.61}                                                       & 93.65                                                   & \phantom{0}74                                                                             & {0.18}      & 84.84                                                   & \phantom{0}526                                                                             & 2.58                                                  & 97.80                                                    & 265                                                                            & 26.57                                                                          \\ \bottomrule
\end{tabular}}
  \label{tab:results2}
  }
    \end{minipage}

}}
\end{table}

\textbf{Baseline algorithms and Implementation}: \salg and \alg were compared to standard RNN~\citep{rumelhart1986}, leading unitary RNN approaches such as SpectralRNN~\citep{zhang2018}, Orthogonal RNN (oRNN)~\citep{mhammedi2016}, Efficient Unitary Recurrent Neural Networks (EURNN)~\citep{jing2016}, FactoredRNN~\citep{vorontsov2017} and state-of-the-art gated RNNs including UGRNN~\citep{collins2016}, GRU~\citep{cho2014} and LSTM~\citep{hochreiter1997}. Details of these methods are provided in Section~\ref{sec:rw}. Native Tensorflow implementations were used for the LSTM and GRU  architectures. For all the other RNNs, publicly available implementations provided by the authors were used taking care to ensure that published results could be reproduced thereby verifying the code and hyper-parameter settings. All experiments were run on an Nvidia Tesla P40 GPU with CUDA 9.0 and cuDNN 7.1 on a machine with an Intel Xeon 2.60 GHz CPU with 12 cores.

\textbf{Hyper-parameters}: The hyper-parameters of each algorithm were set by a fine-grained validation wherever possible or according to the settings recommended by the authors otherwise. Adam, Nesterov Momentum and SGD were used to optimize each algorithm on each dataset and the optimizer with the best validation performance was selected. The learning rate was initialized to $10^{-2}$ for all architectures except for RNNs where the learning rate was initialized to $10^{-3}$ to ensure stable training. Each algorithm was run for 200 epochs after which the learning rate was decreased by a factor of $10^{-1}$ and the algorithm run again for another 100 epochs. This procedure was carried out on all datasets except for Pixel MNIST where the learning rate was decayed by $\half$ after each pass of 200 epochs. Batch sizes between 64 and 128 training points were tried for most architectures and a batch size of 100 was found to work well in general except for standard RNNs which required a batch size of 512. \salg used $\tanh$ as the non-linearity in most cases except for a few (indicated by~\textsuperscript{+}) where ReLU gave slightly better results. Table~\ref{tab:hyp} in the Appendix lists the non-linearity, optimizer and hyper-parameter settings for \alg on all datasets.

\textbf{Evaluation criteria}: The emphasis in this paper is on designing RNN architectures which can run on low-memory IoT devices and which are efficient at prediction time. As such, the model size of each architecture is reported along with its training time and classification accuracy (F1 score on the Wakeword-2 dataset and perplexity on the PTB dataset). Prediction times on some of the popular IoT boards are also reported. Note that, for NLP applications such as PTB and Yelp, just the model size of the various RNN architectures has been reported.  In a real application, the size of the learnt word-vector embeddings (10 MB for \salg and \alg) would also have to be considered.

\textbf{Results}: Tables~\ref{tab:ptb} and~\ref{tab:results1} compare the performance of \salg, \alg and \algfloat to state-of-the-art RNNs. Three points are worth noting about \salg's performance.  First, \salg's prediction accuracy gains over a standard RNN ranged from 2.34\% on the Pixel-MNIST dataset to 19\% on the Google-12 dataset. Second, \salg's prediction accuracy could surpass leading unitary RNNs on 6 out of the 8 datasets with gains up to 2.87\% and 3.77\% over SpectralRNN on the Google-12 and DSA-19 datasets respectively. Third, \salg's training speedups over all unitary and gated RNNs could range from 1.2x over UGRNN on the Yelp-5 and DSA-19 datasets to  196x over EURNN on the Google-12 dataset. This demonstrates that the vanishing and exploding gradient problem could be overcome by the addition of a simple weighted residual connection to the standard RNN architecture thereby allowing \salg to train efficiently and stablely. This also demonstrates that the residual connection offers a theoretically principled architecture that can often result in accuracy gains without limiting the expressive power of the hidden state transition matrix.

Tables~\ref{tab:ptb} and~\ref{tab:results1} also demonstrate that \algfloat could be more accurate and faster to train than all unitary RNNs. Furthermore, \algfloat could match the accuracies and training times of state-of-the-art gated RNNs while having models that could be 1.18-4.87x smaller. This demonstrates that extending the residual connection to a gate which reuses the RNN matrices increased accuracy with virtually no increase in model size over \salg in most cases. In fact, on Google-30 and Pixel-MNIST \algfloat's model size was lower than \salg's as it had a lower hidden dimension indicating that the gate efficiently increased expressive power.

Finally, Tables~\ref{tab:ptb} and~\ref{tab:results1} show that \alg's accuracy was at most 1.13\% worse than the best RNN but its model could be up to 35x smaller even as compared to low-rank unitary methods such as SpectralRNN. Figures~\ref{fig:G12} and~\ref{fig:G30} in the Appendix also show that \algfloat and \alg's classification accuracies could be higher than those obtained by the best unitary and gated RNNs for any given model size in the 0-128 KB range. This demonstrates the effectiveness of making \alg's parameters low-rank, sparse and quantized and allows \alg to fit on the Arduino Uno having just 2 KB RAM and 32 KB flash memory. In particular, \alg was able to recognize the "Hey Cortana" wakeword just as accurately as leading RNNs but with a 1 KB model.

\textbf{Prediction on IoT boards}: Unfortunately, most RNNs were too large to fit on an Arduino Uno apart from \alg. On the slightly more powerful Arduino MKR1000 having an ARM Cortex M0+ microcontroller operating at 48 MHz with 32 KB RAM and 256 KB flash memory, Table~\ref{tab:mkr1000} shows that \alg could achieve the same prediction accuracy while being 25-45x faster at prediction than UGRNN and 57-132x faster than SpectralRNN. Results on the even more powerful Arduino Due are presented in Table~\ref{tab:due} while results on the Raspberry Pi are presented in Table~\ref{tab:rpi} of the Appendix.

\begin{table}[t]
\centering
\vspace{-3mm}
\makebox[0pt][c]{\parbox{1\textwidth}{%

    \begin{minipage}[b]{0.51\hsize}\centering
    \captionsetup{font=small}
     \caption{Prediction time in ms on the Arduino MKR1000}
  \resizebox{0.98\linewidth}{!}
  {
\begin{tabular}{@{}lccc@{}}
\toprule
\multicolumn{1}{l}{Method}  &
\multicolumn{1}{c}{\begin{tabular}[c]{@{}c@{}} Google-12 \end{tabular}}                    & \multicolumn{1}{c}{\begin{tabular}[c]{@{}c@{}}HAR-2\end{tabular}}                    & \multicolumn{1}{c}{\begin{tabular}[c]{@{}c@{}}Wakeword-2\end{tabular}} \\ \midrule

\multirow{1}{*}{\alg}    & \phantom{00}537 &\phantom{0}162 &\phantom{00}175 \\
\multirow{1}{*}{\alg-Q}    & \phantom{0}2282 &\phantom{0}553 &\phantom{00}755 \\
\multirow{1}{*}{RNN}    & 12028 & 2249 &\phantom{0}2232 \\
\multirow{1}{*}{UGRNN}    & 22875 & 4207 &\phantom{0}6724 \\
\multirow{1}{*}{SpectralRNN}    & 70902 & --- &10144 \\ \bottomrule
\end{tabular}

\label{tab:mkr1000}
  }
    \end{minipage}
    \hfill
    \begin{minipage}[b]{0.51\hsize}\centering
    \captionsetup{font=small}
       \caption{Prediction time in ms on the Arduino Due}

  \resizebox{0.98\linewidth}{!}
  {
\begin{tabular}{@{}lccc@{}}
\toprule
\multicolumn{1}{l}{Method}  &
\multicolumn{1}{c}{\begin{tabular}[c]{@{}c@{}} Google-12 \end{tabular}}                    & \multicolumn{1}{c}{\begin{tabular}[c]{@{}c@{}}HAR-2\end{tabular}}                    & \multicolumn{1}{c}{\begin{tabular}[c]{@{}c@{}}Wakeword-2\end{tabular}} \\ \midrule
\multirow{1}{*}{\alg}    & \phantom{00}242 &\phantom{000}62 &\phantom{00}77 \\
\multirow{1}{*}{\alg-Q}    & \phantom{00}779 &\phantom{00}172 &\phantom{0}238 \\
\multirow{1}{*}{RNN}    & \phantom{0}3472 & \phantom{00}590 &\phantom{0}653 \\
\multirow{1}{*}{UGRNN}    & \phantom{0}6693 & \phantom{0}1142 &1823 \\
\multirow{1}{*}{SpectralRNN}    & 17766 & 55558 &2691 \\ \bottomrule
\end{tabular}

  \label{tab:due}
  }
    \end{minipage}
}}
\vspace{-2mm}
\end{table}

\textbf{Ablations, extensions and parameter settings}:
Enforcing that \alg's matrices be low-rank led to a slight increase in prediction accuracy and reduction in prediction costs as shown in the ablation experiments in Tables~\ref{tab:cRes},~\ref{tab:cWake} and ~\ref{tab:cPTB} in the Appendix. Adding sparsity and quantization led to a slight drop in accuracy but resulted in significantly smaller models. Next, Table~\ref{tab:PTBReg} in the Appendix shows that regularization and layering techniques~\citep{merity2017regularizing} that have been proposed to increase the prediction accuracy of other gated RNNs are also effective for \alg and can lead to reductions in perplexity on the PTB dataset. Finally, Figure~\ref{fig:plt_ab} and Table~\ref{tab:alphaT} of the Appendix measure the agreement between \salg's theoretical analysis and empirical observations. Figure~\ref{fig:plt_ab} (a) shows that the $\alpha$ learnt on datasets with $T$ time steps is decreasing function of $T$ and Figure~\ref{fig:plt_ab} (b) shows that the learnt $\alpha$ and $\beta$ follow the relation $\alpha/\beta\approx O(1/T)$ for large $T$ which is one of the settings in which \salg's gradients stabilize and training converges quickly as proved by Theorems~\ref{thm:conv} and~\ref{thm:gen}. Furthermore, $\beta$ can be seen to be close to $1-\alpha$ for large $T$ in Figure~\ref{fig:plt_ab} (c) as assumed in Section~\ref{sec:analysis} for the convergence of long sequences. For instance, the relative error between $\beta$ and $1-\alpha$  for Google-12 with 99 timesteps was 2.15\%, for HAR-2 with 128 timesteps was 3.21\% and for MNIST-10 with 112 timesteps was 0.68\%. However, for short sequences where there was a lower likelihood of gradients exploding or vanishing, $\beta$ was found to deviate significantly from $1-\alpha$ as this led to improved prediction accuracy. Enforcing that $\beta = 1-\alpha$ on short sequences was found to drop accuracy by up to 1.5\%.
\vspace{-1mm}
\begin{figure*}[h!]
\centering\hspace*{-3ex}
\vspace{-1mm}
\begin{tabular}{ccc}
  \includegraphics[width=.35\textwidth]{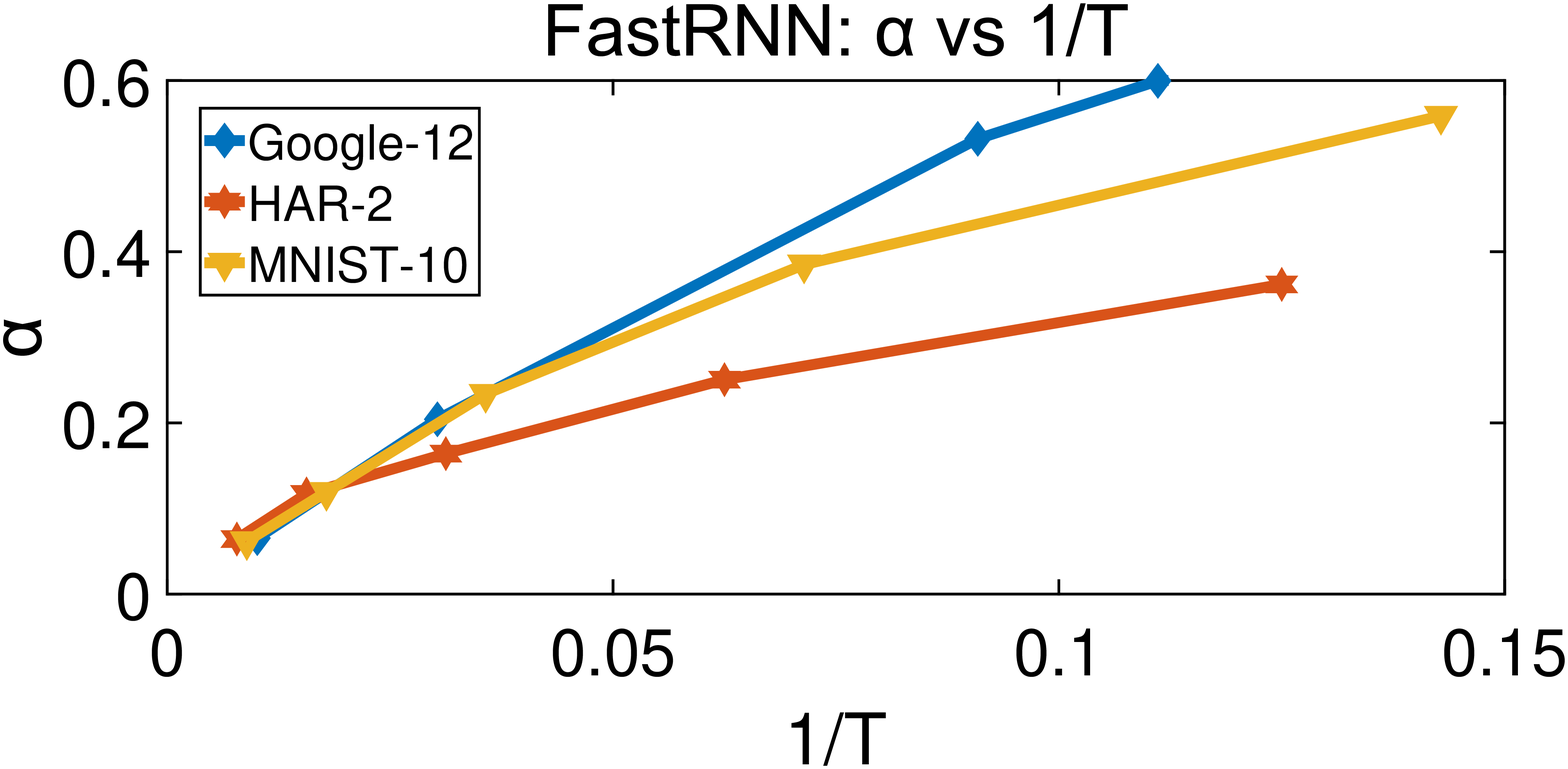}&
  \hspace{-4.25ex}
  \includegraphics[width=.35\textwidth]{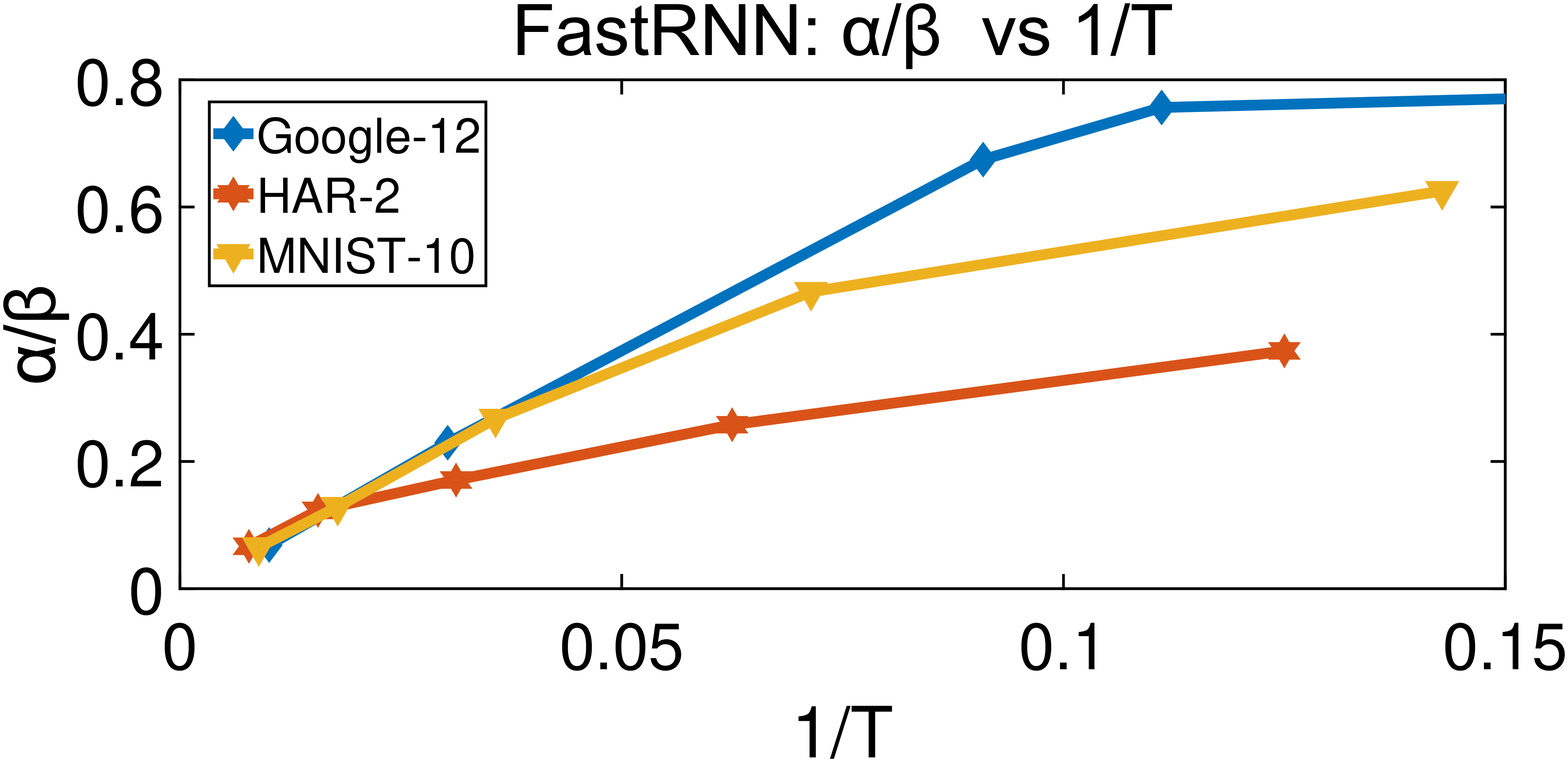}&
  \hspace{-4.5ex}
  \includegraphics[width=.35\textwidth]{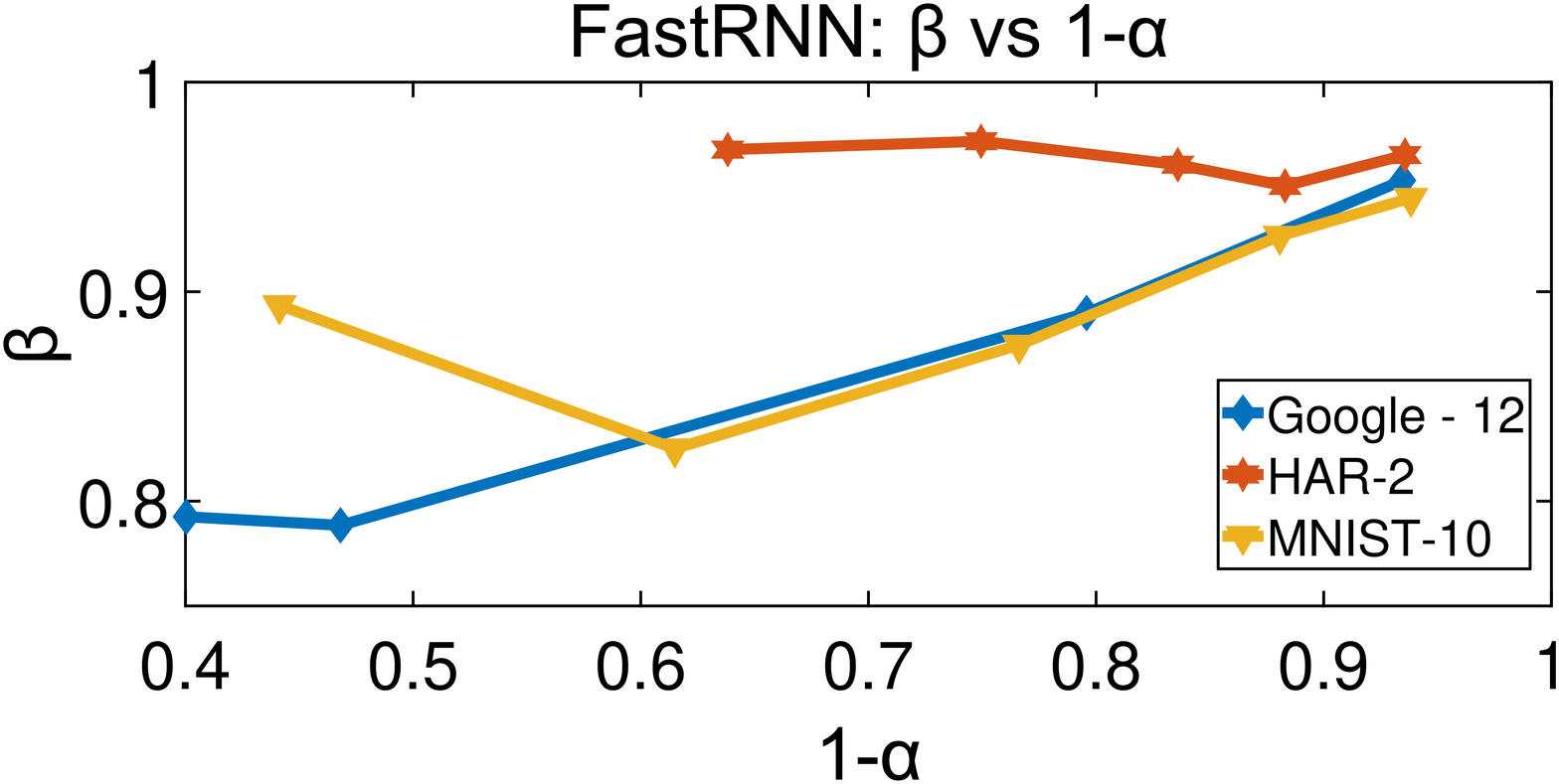}\\
(a)&(b)&(c)
\end{tabular}
\caption{\small{Plots (a) and (b) show the variation of $\alpha$ and $\nicefrac{\alpha}{\beta}$ of \salg with respect to $\nicefrac{1}{T}$ for three datasets. Plot (c) shows the relation between $\beta$ and $1-\alpha$. In accordance with Theorem \ref{thm:conv}, the learnt values of $\alpha$ and $\alpha/\beta$ scale as $O(1/T)$ while $\beta \rightarrow 1-\alpha$ for long sequences. }}
  \label{fig:plt_ab}
  \vspace{-6mm}
  \vspace{1mm}
\end{figure*}

\section{Conclusions}
\label{sec:conc}
This paper proposed the \salg and \alg architectures for efficient RNN training and prediction. \salg could lead to provably stable training by incorporating a residual connection with two scalar parameters into the standard RNN architecture. \salg was demonstrated to have lower training times, lower prediction costs and higher prediction accuracies than leading unitary RNNs in most cases. \alg extended the residual connection to a gate reusing the RNN matrices and was able to match the accuracies of state-of-the-art gated RNNs but with significantly lower prediction costs. \alg's model could be compressed to 1-6 KB without compromising accuracy in many cases by enforcing that its parameters be low-rank, sparse and quantized. This allowed \alg to make accurate predictions efficiently on severely resource-constrained IoT devices too tiny to hold other RNN models.

\section*{Acknowledgements}
\label{sec:ack}

We are grateful to Ankit Anand, Niladri Chatterji, Kunal Dahiya, Don Dennis, Inderjit S. Dhillon, Dinesh Khandelwal, Shishir Patil, Adithya Pratapa, Harsha Vardhan Simhadri and Raghav Somani for helpful discussions and feedback. KB acknowledges the support of the NSF through grant IIS-1619362 and of the AFOSR through grant FA9550-17-1-0308.

\bibliography{local}

\begin{thebibliography}{54}
\providecommand{\natexlab}[1]{#1}
\providecommand{\url}[1]{\texttt{#1}}
\expandafter\ifx\csname urlstyle\endcsname\relax
  \providecommand{\doi}[1]{doi: #1}\else
  \providecommand{\doi}{doi: \begingroup \urlstyle{rm}\Url}\fi

\bibitem[Ahmad et~al.(2017)Ahmad, Lavin, Purdy, and
  Agha]{ahmad2017unsupervised}
S.~Ahmad, A.~Lavin, S.~Purdy, and Z.~Agha.
\newblock Unsupervised real-time anomaly detection for streaming data.
\newblock \emph{Neurocomputing}, 262:\penalty0 134--147, 2017.

\bibitem[Altun et~al.(2010)Altun, Barshan, and
  Tun{\c{c}}el]{altun2010comparative}
K.~Altun, B.~Barshan, and O.~Tun{\c{c}}el.
\newblock Comparative study on classifying human activities with miniature
  inertial and magnetic sensors.
\newblock \emph{Pattern Recognition}, 43\penalty0 (10):\penalty0 3605--3620,
  2010.
\newblock URL
  \url{https://archive.ics.uci.edu/ml/datasets/Daily+and+Sports+Activities}.

\bibitem[Anguita et~al.(2012)Anguita, Ghio, Oneto, Parra, and
  Reyes-Ortiz]{anguita2012}
D.~Anguita, A.~Ghio, L.~Oneto, X.~Parra, and J.~L. Reyes-Ortiz.
\newblock Human activity recognition on smartphones using a multiclass
  hardware-friendly support vector machine.
\newblock In \emph{International Workshop on Ambient Assisted Living}, pages
  216--223. Springer, 2012.
\newblock URL
  \url{https://archive.ics.uci.edu/ml/datasets/human+activity+recognition+using+smartphones}.

\bibitem[Anthony and Bartlett(2009)]{anthony2009neural}
M.~Anthony and P.~L. Bartlett.
\newblock \emph{Neural network learning: Theoretical foundations}.
\newblock Cambridge University Press, 2009.

\bibitem[Arjovsky et~al.(2016)Arjovsky, Shah, and Bengio]{arjovsky2016}
M.~Arjovsky, A.~Shah, and Y.~Bengio.
\newblock Unitary evolution recurrent neural networks.
\newblock In \emph{International Conference on Machine Learning}, pages
  1120--1128, 2016.

\bibitem[Bartlett and Mendelson(2002)]{bartlett2002}
P.~L. Bartlett and S.~Mendelson.
\newblock Rademacher and gaussian complexities: Risk bounds and structural
  results.
\newblock \emph{Journal of Machine Learning Research}, 3\penalty0
  (Nov):\penalty0 463--482, 2002.

\bibitem[Bengio et~al.(2013)Bengio, Boulanger-Lewandowski, and
  Pascanu]{bengio2013}
Y.~Bengio, N.~Boulanger-Lewandowski, and R.~Pascanu.
\newblock Advances in optimizing recurrent networks.
\newblock In \emph{Acoustics, Speech and Signal Processing (ICASSP), 2013 IEEE
  International Conference on}, pages 8624--8628. IEEE, 2013.

\bibitem[Bhatia et~al.()Bhatia, Dahiya, Jain, Prabhu, and Varma]{xmlcode}
K.~Bhatia, K.~Dahiya, H.~Jain, Y.~Prabhu, and M.~Varma.
\newblock {The Extreme Classification Repository: Multi-label Datasets} \&
  {Code}.
\newblock URL \url{http://manikvarma.org/downloads/XC/XMLRepository.html}.

\bibitem[Bradbury et~al.(2016)Bradbury, Merity, Xiong, and
  Socher]{bradbury2016quasi}
J.~Bradbury, S.~Merity, C.~Xiong, and R.~Socher.
\newblock Quasi-recurrent neural networks.
\newblock \emph{arXiv preprint arXiv:1611.01576}, 2016.

\bibitem[Campos et~al.(2018)Campos, Jou, i~Nieto, Torres, and
  Chang]{campos2018}
V.~Campos, B.~Jou, X.~G. i~Nieto, J.~Torres, and S.-F. Chang.
\newblock Skip {RNN}: Learning to skip state updates in recurrent neural
  networks.
\newblock In \emph{International Conference on Learning Representations}, 2018.

\bibitem[Chen et~al.(2014)Chen, Parada, and Heigold]{chen2014small}
G.~Chen, C.~Parada, and G.~Heigold.
\newblock Small-footprint keyword spotting using deep neural networks.
\newblock In \emph{Acoustics, Speech and Signal Processing (ICASSP), 2015 IEEE
  International Conference on}, pages 4087--4091. IEEE, 2014.

\bibitem[Chen et~al.(2015)Chen, Parada, and Sainath]{chen2015query}
G.~Chen, C.~Parada, and T.~N. Sainath.
\newblock Query-by-example keyword spotting using long short-term memory
  networks.
\newblock In \emph{Acoustics, Speech and Signal Processing (ICASSP), 2015 IEEE
  International Conference on}, pages 5236--5240. IEEE, 2015.

\bibitem[Cho et~al.(2014)Cho, Van~Merri{\"e}nboer, Bahdanau, and
  Bengio]{cho2014}
K.~Cho, B.~Van~Merri{\"e}nboer, D.~Bahdanau, and Y.~Bengio.
\newblock On the properties of neural machine translation: Encoder-decoder
  approaches.
\newblock \emph{arXiv preprint arXiv:1409.1259}, 2014.

\bibitem[Collins et~al.(2016)Collins, Sohl-Dickstein, and
  Sussillo]{collins2016}
J.~Collins, J.~Sohl-Dickstein, and D.~Sussillo.
\newblock Capacity and trainability in recurrent neural networks.
\newblock \emph{arXiv preprint arXiv:1611.09913}, 2016.

\bibitem[Ghadimi and Lan(2013)]{ghadimi2013}
S.~Ghadimi and G.~Lan.
\newblock Stochastic first-and zeroth-order methods for nonconvex stochastic
  programming.
\newblock \emph{SIAM Journal on Optimization}, 23\penalty0 (4):\penalty0
  2341--2368, 2013.

\bibitem[Golowich et~al.(2017)Golowich, Rakhlin, and Shamir]{golowich2017}
N.~Golowich, A.~Rakhlin, and O.~Shamir.
\newblock Size-independent sample complexity of neural networks.
\newblock \emph{arXiv preprint arXiv:1712.06541}, 2017.

\bibitem[Gupta et~al.(2017)Gupta, Suggala, Gupta, Simhadri, Paranjape, Kumar,
  Goyal, Udupa, Varma, and Jain]{Gupta17}
C.~Gupta, A.~S. Suggala, A.~Gupta, H.~V. Simhadri, B.~Paranjape, A.~Kumar,
  S.~Goyal, R.~Udupa, M.~Varma, and P.~Jain.
\newblock Protonn: Compressed and accurate knn for resource-scarce devices.
\newblock In \emph{Proceedings of the International Conference on Machine
  Learning}, August 2017.

\bibitem[Han et~al.(2016)Han, Mao, and Dally]{han16}
S.~Han, H.~Mao, and W.~J. Dally.
\newblock Deep compression: Compressing deep neural networks with pruning,
  trained quantization and huffman coding.
\newblock In \emph{ICLR}, 2016.

\bibitem[He et~al.(2016)He, Zhang, Ren, and Sun]{he2016}
K.~He, X.~Zhang, S.~Ren, and J.~Sun.
\newblock Deep residual learning for image recognition.
\newblock In \emph{Proceedings of the IEEE Conference on Computer Vision and
  Pattern Recognition}, pages 770--778, 2016.

\bibitem[Hochreiter and Schmidhuber(1997)]{hochreiter1997}
S.~Hochreiter and J.~Schmidhuber.
\newblock Long short-term memory.
\newblock \emph{Neural computation}, 9\penalty0 (8):\penalty0 1735--1780, 1997.

\bibitem[Inan et~al.(2016)Inan, Khosravi, and Socher]{inan2016tying}
H.~Inan, K.~Khosravi, and R.~Socher.
\newblock Tying word vectors and word classifiers: A loss framework for
  language modeling.
\newblock \emph{arXiv preprint arXiv:1611.01462}, 2016.

\bibitem[Jaeger et~al.(2007)Jaeger, Luko{{s}}evi{{c}}ius, Popovici, and
  Siewert]{jaeger2007}
H.~Jaeger, M.~Luko{{s}}evi{{c}}ius, D.~Popovici, and U.~Siewert.
\newblock Optimization and applications of echo state networks with
  leaky-integrator neurons.
\newblock \emph{Neural Networks}, 20\penalty0 (3):\penalty0 335--352, 2007.

\bibitem[Jing et~al.(2017{\natexlab{a}})Jing, Gulcehre, Peurifoy, Shen,
  Tegmark, Solja{{c}}i{\'c}, and Bengio]{jing2017}
L.~Jing, C.~Gulcehre, J.~Peurifoy, Y.~Shen, M.~Tegmark, M.~Solja{{c}}i{\'c},
  and Y.~Bengio.
\newblock Gated orthogonal recurrent units: On learning to forget.
\newblock \emph{arXiv preprint arXiv:1706.02761}, 2017{\natexlab{a}}.

\bibitem[Jing et~al.(2017{\natexlab{b}})Jing, Shen, Dub{{c}}ek, Peurifoy,
  Skirlo, Tegmark, and Solja{{c}}i{\'c}]{jing2016}
L.~Jing, Y.~Shen, T.~Dub{{c}}ek, J.~Peurifoy, S.~Skirlo, M.~Tegmark, and
  M.~Solja{{c}}i{\'c}.
\newblock Tunable efficient unitary neural networks (eunn) and their
  application to {RNN}.
\newblock In \emph{International Conference on Machine Learning},
  2017{\natexlab{b}}.

\bibitem[Jose et~al.(2018)Jose, Cisse, and Fleuret]{jose2017}
C.~Jose, M.~Cisse, and F.~Fleuret.
\newblock {K}ronecker recurrent units.
\newblock In J.~Dy and A.~Krause, editors, \emph{International Conference on
  Machine Learning}, volume~80 of \emph{Proceedings of Machine Learning
  Research}, pages 2380--2389, Stockholmsmässan, Stockholm Sweden, 10--15 Jul
  2018. PMLR.

\bibitem[Kanai et~al.(2017)Kanai, Fujiwara, and Iwamura]{kanai2017}
S.~Kanai, Y.~Fujiwara, and S.~Iwamura.
\newblock Preventing gradient explosions in gated recurrent units.
\newblock In \emph{Advances in Neural Information Processing Systems}, pages
  435--444, 2017.

\bibitem[K{\"e}puska and Klein(2009)]{kepuska2009novel}
V.~K{\"e}puska and T.~Klein.
\newblock A novel wake-up-word speech recognition system, wake-up-word
  recognition task, technology and evaluation.
\newblock \emph{Nonlinear Analysis: Theory, Methods \& Applications},
  71\penalty0 (12):\penalty0 e2772--e2789, 2009.

\bibitem[Kingma and Ba(2014)]{kingma2014}
D.~P. Kingma and J.~Ba.
\newblock Adam: A method for stochastic optimization.
\newblock \emph{arXiv preprint arXiv:1412.6980}, 2014.

\bibitem[Kumar et~al.(2017)Kumar, Goyal, and Varma]{Kumar17}
A.~Kumar, S.~Goyal, and M.~Varma.
\newblock Resource-efficient machine learning in 2 kb ram for the internet of
  things.
\newblock In \emph{Proceedings of the International Conference on Machine
  Learning}, August 2017.

\bibitem[Kusupati et~al.(2017)Kusupati, Dennis, Gupta, Kumar, Patil, and
  Simhadri]{edgemlcode}
A.~Kusupati, D.~Dennis, C.~Gupta, A.~Kumar, S.~Patil, and H.~Simhadri.
\newblock {The EdgeML Library: An ML library for machine learning on the Edge},
  2017.
\newblock URL \url{https://github.com/Microsoft/EdgeML}.

\bibitem[Le et~al.(2015)Le, Jaitly, and Hinton]{le2015simple}
Q.~V. Le, N.~Jaitly, and G.~E. Hinton.
\newblock A simple way to initialize recurrent networks of rectified linear
  units.
\newblock \emph{arXiv preprint arXiv:1504.00941}, 2015.

\bibitem[LeCun et~al.(1998)LeCun, Bottou, Bengio, and Haffner]{lecun1998}
Y.~LeCun, L.~Bottou, Y.~Bengio, and P.~Haffner.
\newblock Gradient-based learning applied to document recognition.
\newblock \emph{Proceedings of the IEEE}, 86\penalty0 (11):\penalty0
  2278--2324, 1998.

\bibitem[Marcus et~al.(1993)Marcus, Marcinkiewicz, and
  Santorini]{marcus1993building}
M.~P. Marcus, M.~A. Marcinkiewicz, and B.~Santorini.
\newblock Building a large annotated corpus of english: The penn treebank.
\newblock \emph{Computational linguistics}, 19\penalty0 (2):\penalty0 313--330,
  1993.

\bibitem[McAuley and Leskovec(2013)]{mcauley2013hidden}
J.~McAuley and J.~Leskovec.
\newblock Hidden factors and hidden topics: understanding rating dimensions
  with review text.
\newblock In \emph{Proceedings of the 7th ACM conference on Recommender
  systems}, pages 165--172. ACM, 2013.

\bibitem[Melis et~al.(2017)Melis, Dyer, and Blunsom]{melis2017state}
G.~Melis, C.~Dyer, and P.~Blunsom.
\newblock On the state of the art of evaluation in neural language models.
\newblock \emph{arXiv preprint arXiv:1707.05589}, 2017.

\bibitem[Merity et~al.(2017)Merity, Keskar, and Socher]{merity2017regularizing}
S.~Merity, N.~S. Keskar, and R.~Socher.
\newblock Regularizing and optimizing {LSTM} language models.
\newblock \emph{arXiv preprint arXiv:1708.02182}, 2017.

\bibitem[Mhammedi et~al.(2017)Mhammedi, Hellicar, Rahman, and
  Bailey]{mhammedi2016}
Z.~Mhammedi, A.~Hellicar, A.~Rahman, and J.~Bailey.
\newblock Efficient orthogonal parametrisation of recurrent neural networks
  using householder reflections.
\newblock In \emph{International Conference on Machine Learning}, 2017.

\bibitem[Mikolov and Zweig(2012)]{mikolov2012context}
T.~Mikolov and G.~Zweig.
\newblock Context dependent recurrent neural network language model.
\newblock \emph{SLT}, 12\penalty0 (234-239):\penalty0 8, 2012.

\bibitem[Narang et~al.(2017)Narang, Elsen, Diamos, and Sengupta]{narang2017}
S.~Narang, E.~Elsen, G.~Diamos, and S.~Sengupta.
\newblock Exploring sparsity in recurrent neural networks.
\newblock \emph{arXiv preprint arXiv:1704.05119}, 2017.

\bibitem[Pascanu et~al.(2013)Pascanu, Mikolov, and Bengio]{pascanu2013}
R.~Pascanu, T.~Mikolov, and Y.~Bengio.
\newblock On the difficulty of training recurrent neural networks.
\newblock In \emph{International Conference on Machine Learning}, pages
  1310--1318, 2013.

\bibitem[Rumelhart et~al.(1986)Rumelhart, Hinton, and Williams]{rumelhart1986}
D.~E. Rumelhart, G.~E. Hinton, and R.~J. Williams.
\newblock Learning representations by back-propagating errors.
\newblock \emph{Nature}, 323\penalty0 (6088):\penalty0 533, 1986.

\bibitem[Sainath and Parada(2015)]{sainath2015convolutional}
T.~N. Sainath and C.~Parada.
\newblock Convolutional neural networks for small-footprint keyword spotting.
\newblock In \emph{Sixteenth Annual Conference of the International Speech
  Communication Association}, 2015.

\bibitem[{Siri Team, Apple}(2017)]{siri}
{Siri Team, Apple}.
\newblock Hey {Siri}: An on-device dnn-powered voice trigger for apple’s
  personal assistant, 2017.
\newblock URL \url{https://machinelearning.apple.com/2017/10/01/hey-siri.html}.

\bibitem[Srivastava et~al.(2015)Srivastava, Greff, and
  Schmidhuber]{srivastava2015}
R.~K. Srivastava, K.~Greff, and J.~Schmidhuber.
\newblock Highway networks.
\newblock \emph{arXiv preprint arXiv:1505.00387}, 2015.

\bibitem[{STCI, Microsoft}()]{wakeword}
{STCI, Microsoft}.
\newblock Wakeword dataset.

\bibitem[Susto et~al.(2015)Susto, Schirru, Pampuri, McLoone, and
  Beghi]{susto2015machine}
G.~A. Susto, A.~Schirru, S.~Pampuri, S.~McLoone, and A.~Beghi.
\newblock Machine learning for predictive maintenance: A multiple classifier
  approach.
\newblock \emph{IEEE Transactions on Industrial Informatics}, 11\penalty0
  (3):\penalty0 812--820, 2015.

\bibitem[Vorontsov et~al.(2017)Vorontsov, Trabelsi, Kadoury, and
  Pal]{vorontsov2017}
E.~Vorontsov, C.~Trabelsi, S.~Kadoury, and C.~Pal.
\newblock On orthogonality and learning recurrent networks with long term
  dependencies.
\newblock In \emph{International Conference on Machine Learning}, 2017.

\bibitem[Wang et~al.(2017)Wang, Lin, and Wang]{wang2017accelerating}
Z.~Wang, J.~Lin, and Z.~Wang.
\newblock Accelerating recurrent neural networks: A memory-efficient approach.
\newblock \emph{IEEE Transactions on Very Large Scale Integration (VLSI)
  Systems}, 25\penalty0 (10):\penalty0 2763--2775, 2017.

\bibitem[Warden(2018)]{warden2017}
P.~Warden.
\newblock Speech commands: A dataset for limited-vocabulary speech recognition.
\newblock \emph{arXiv preprint arXiv:1804.03209}, 2018.
\newblock URL
  \url{http://download.tensorflow.org/data/speech_commands_v0.01.tar.gz}.

\bibitem[Wisdom et~al.(2016)Wisdom, Powers, Hershey, Le~Roux, and
  Atlas]{wisdom2016}
S.~Wisdom, T.~Powers, J.~Hershey, J.~Le~Roux, and L.~Atlas.
\newblock Full-capacity unitary recurrent neural networks.
\newblock In \emph{Advances in Neural Information Processing Systems}, pages
  4880--4888, 2016.

\bibitem[Ye et~al.(2017)Ye, Wang, Li, Chen, Zhe, Chu, and Xu]{ye2017learning}
J.~Ye, L.~Wang, G.~Li, D.~Chen, S.~Zhe, X.~Chu, and Z.~Xu.
\newblock Learning compact recurrent neural networks with block-term tensor
  decomposition.
\newblock \emph{arXiv preprint arXiv:1712.05134}, 2017.

\bibitem[{Yelp Inc}(2017)]{yelp}
{Yelp Inc}.
\newblock Yelp dataset challenge, 2017.
\newblock URL \url{https://www.yelp.com/dataset/challenge}.

\bibitem[Zaremba et~al.(2014)Zaremba, Sutskever, and
  Vinyals]{zaremba2014recurrent}
W.~Zaremba, I.~Sutskever, and O.~Vinyals.
\newblock Recurrent neural network regularization.
\newblock \emph{arXiv preprint arXiv:1409.2329}, 2014.

\bibitem[Zhang et~al.(2018)Zhang, Lei, and Dhillon]{zhang2018}
J.~Zhang, Q.~Lei, and I.~S. Dhillon.
\newblock Stabilizing gradients for deep neural networks via efficient {SVD}
  parameterization.
\newblock In \emph{International Conference on Machine Learning}, 2018.

\end{thebibliography}
\newpage
\appendix
\section{Convergence Analysis for \salg}
\label{sec:conv}
\begin{algorithm}[ht!]

  \DontPrintSemicolon
  \KwIn{Initial point $\v \theta_1$, iteration limit M, step sizes $\gamma_{k\geq 1}$, Probability mass function $P_R(\cdot)$ supported on $\{1, 2, \ldots, M \}$}
    \textbf{Initialize:} $R$ be a random variable with probability mass function $P_R$\;
  \For{$m = 1, \ldots, R$}{
  Obtain sample of stochastic gradient $\nabla L_t(\v \theta_t)$\;
  $\v \theta_t \leftarrow \v \theta_{t-1} - \gamma_t \nabla L_t(\v \theta_t)$\;
  }
  \KwOut{$\v \theta_R$}
  \caption{Randomized Stochastic Gradient}
  \label{alg:rsg}
\end{algorithm}

Let $\v \theta = (\vec{W}, \vec{U}, \vec{v})$ represent the set of parameters of the scalar gated recurrent neural network. In order to prove the convergence properties of Randomized Stochastic Gradient (see Algorithm \ref{alg:rsg}) as in \cite{ghadimi2013}, we first obtain a bound on the Lipschitz constant of the loss function $L(\vec{X}, y;\theta) :=  \log(1+\exp{(-y\cdot\vec{v}^\top \vec{h}_T)})$ where $\vec{h}_T$ is the output of \salg after $T$ time steps given input $\vec{X}$.

The gradient $\nabla_\theta L$ of the loss function is given by $(\frac{\partial L}{\partial \vec{W}}, \frac{\partial L}{\partial \vec{U}}, \frac{\partial L}{\partial \vec{v}})$ wherein
\begin{small}
\begin{align}
\frac{\partial L}{\partial \vec{U}} &= \alpha \sum_{t=0}^{T} \vec{D}_t \left(\prod_{k = t}^{T-1}(\alpha \vec{U}^\top \vec{D}_{k+1} + \beta \vec{I}) \right)(\nabla_{\vec{h}_T}L)\vec{h}_{t-1}^\top\\
\frac{\partial L}{\partial \vec{W}} &= \alpha \sum_{t=0}^{T} \vec{D}_t \left(\prod_{k = t}^{T-1}(\alpha \vec{U}^\top \vec{D}_{k+1} + \beta \vec{I}) \right)(\nabla_{\vec{h}_T}L) \v x_{t}^\top\\
\frac{\partial L}{\partial \vec{v}} &= \frac{-y\exp{(-y\cdot\vec{v} ^\top \vec{h}_T)}}{1+\exp{(-y\cdot\vec{v} ^\top \vec{h}_T)}} \vec{h}_T,
\end{align}
\end{small}

where $\nabla_{\vec{h}_T}L = -c(\theta)y\cdot \vec{v}$, with $c(\theta) = \frac{1}{1+\exp{(y\cdot\vec{v}^\top \vec{h}_T)}}$. We do a perturbation analysis and obtain a bound on $\|\nabla _{\theta}L(\theta) - \nabla _{\theta}L(\theta + \delta) \|_2$ where $\delta = (\delta_{\vec{W}}, \delta_{\vec{U}}, \delta_{\vec{v}})$.

\textbf{Deviation bound for $\vec{h}_T$}: In this subsection, we consider bounding the term $\| \vec{h}_T(\theta + \delta) - \vec{h}_T(\theta)\|_2$ evaluated on the same input $\vec{X}$. Note that for \salg, $\vec{h}_T = \alpha \vec{\tilde{h}}_T + \beta \vec{h}_{T-1}$. For notational convenience, we use $\vec{h}_T' = \vec{h}_T(\theta+\delta)$ and $\vec{h}_T = \vec{h}_T(\theta)$.
\begin{small}
\begin{align}\label{eq:bnd_h}
\| \vec{h}_T' - \vec{h}_T \|_2 &\leq \beta\|\vec{h}_{T-1}' - \vec{h}_{T-1} \|_2 + \alpha\|\sigma(\vec{W}\vec{x}_{T} + \vec{U}\vec{h}_{T-1}) - \sigma((\vec{W} + \delta_{\vec{W}})\vec{x}_{T} + (\vec{U}+\delta_{\vec{U}})\vec{h}_{T-1}')\|_2\nonumber\\
&\stackrel{\zeta_1}{\leq} \beta\|\vec{h}_{T-1}' - \vec{h}_{T-1} \|_2 +\alpha  \| \vec{U}\vec{h}_{T-1} - \delta_{\vec{W}}\vec{x}_{T} - \vec{U}\vec{h}_{T-1}' - \delta_{\vec{U}}\vec{h}_{T-1}'\|_2\nonumber\\
&\leq (\alpha  \|\vec{U}\|_2 + \beta)\|\vec{h}_{T-1}' - \vec{h}_{T-1} \|_2  + \alpha (\sqrt{\hat{D}}\cdot \|\delta_{\vec{U}}\|_2 + \| \delta_{\vec{W}}\|_2R_{\v x})\nonumber\\
&\vdots\nonumber\\
&\leq \alpha (\sqrt{\hat{D}}\cdot \|\delta_{\vec{U}} \|_2 + \|\delta_{\vec{W}}\|_2R_{\v x})\left(1+ (\alpha \|\vec{U}\|_2 + \beta) + \ldots + (\alpha  \|\vec{U}\|_2 + \beta)^{T-1} \right)\nonumber\\
&\stackrel{\zeta_2}{\leq} \alpha ( \sqrt{\hat{D}}\cdot \|\delta_{\vec{U}} \|_2 + \|\delta_{\vec{W}}\|_2R_{\v x})\frac{\left(\alpha ( \| \vec{U}\|_2-1) + 1\right)^{T}-1}{\alpha( \| \vec{U}\|_2-1)} \leq 2\alpha( \sqrt{\hat{D}}\cdot  \|\delta_{\vec{U}} \|_2 + \|\delta_{\vec{W}}\|_2R_{\v x})\cdot T\nonumber\\
&\leq \frac{2\sqrt{\hat{D}}\cdot  \|\delta_{\vec{U}} \|_2 + 2\|\delta_{\vec{W}}\|_2R_{\v x}}{|\|\vec{U}\|_2-1|},
\end{align}
\end{small}
where $\zeta_1$ follows by using $1$-lipschitz property of the sigmoid function and $\zeta_2$ follows by setting $\alpha = O(\frac{1}{T\cdot |\|\vec{U}\|_2-1|})$ and $\beta = 1-\alpha$.

\textbf{Deviation bound for $c(\theta)$}: In this subsection, we consider bounding the deviation $c(\theta) - c(\theta+\delta)$.
\begin{small}
\begin{align}
|c(\theta) - c(\theta+\delta)| &\leq |\vec{v}^\top \vec{h}_T - (\vec{v} + \delta_{\vec{v}}^\top)\vec{h}_T' |\nonumber \\
&\leq |\vec{v}^\top (\vec{h}_T-\vec{h}_T')| + \|\delta_{\vec{v}}\|_2\|\vec{h}_t \|_2\nonumber \\
&\leq \|\vec{v}\|_2\|\vec{h}_T -\vec{h}_T'\|_2 + \|\delta_{\vec{v}}\|_2\|\vec{h}_t \|_2\nonumber \\
&\leq R_{\vec{v}}\|\vec{h}_T -\vec{h}_T'\|_2 + \sqrt{\hat{D}}\|\delta_{\vec{v}}\|_2.
\end{align}
\end{small}

\textbf{Deviation bound for $\frac{\partial L}{\partial \vec{v}}$}: In this subsection we consider the bounds on $\|\frac{\partial L}{\partial \vec{v}} (\theta) - \frac{\partial L}{\partial \vec{v}}(\theta+\delta)\|_2$.
\begin{small}
\begin{align}
\left\Vert \frac{\partial L}{\partial \vec{v}} (\theta) - \frac{\partial L}{\partial \vec{v}}(\theta+\delta)\right\Vert_2 &= \left\Vert \frac{ \vec{h}_T}{1+\exp{(y\vec{v}^\top \vec{h}_T)}} - \frac{ \vec{h}_T'}{1+\exp{(y(\vec{v} + \delta_{\vec{v}})^\top \vec{h}_T')}}  \right\Vert_2\nonumber\\
&= \left\Vert c(\theta)  \vec{h}_T - c(\theta + \delta) \vec{h}_T') \right\Vert_2\nonumber\\
&=\|(c(\theta) - c(\theta+\delta))\cdot \vec{h}_T +  c(\theta+\delta)\cdot(\vec{h}_T - \vec{h}_T')\|_2\nonumber\\
&\leq \sqrt{\hat{D}}\cdot |c(\theta) - c(\theta+\delta)| + \|\vec{h}_T - \vec{h}_T'\|_2\nonumber\\
&\leq \left(\sqrt{\hat{D}} R_\vec{v} +1\right)\cdot \|\vec{h}_T -\vec{h}_T'\|_2 + {\hat{D}}\|\delta_{\vec{v}}\|_2.
\end{align}
\end{small}

\textbf{Deviation bound for $\frac{\partial L}{\partial \vec{W}}$}: In this subsection, we analyze $\|\frac{\partial L}{\partial \vec{W}} (\theta) - \frac{\partial L}{\partial \vec{W}} (\theta + \delta)\|_2$. Let $\D= \sup_{k, \theta} \|\vec{D}_k^\theta\|$.
\begin{small}
\begin{align}
&\left\lVert\frac{\partial L}{\partial \vec{W}} (\theta) - \frac{\partial L}{\partial \vec{W}} (\theta + \delta)\right\rVert_F \nonumber\\
&= \alpha R_{\v x}  \left\Vert\sum_{t=0}^T
\left[\left(
c(\theta)\vec{D}_t^\theta \prod_{k = t}^{T-1}(\alpha \vec{U}^\top \vec{D}_{k+1}^\theta + \beta \vec{I})
\right)
\vec{v}
-
\left(
c(\theta+\delta)\vec{D}_t^{\theta+\delta}\prod_{k = t}^{T-1}(\alpha (\vec{U}+\delta_{\vec{U}})^\top \vec{D}_{k+1}^{\theta+\delta} + \beta \vec{I})
\right)
(\vec{v} + \delta_{\vec{v}})
\right]\right\Vert_2.
\label{eq:dl1}
\end{align}
\end{small}

Let us define matrices $\A_t^\theta : = \vec{D}_t^\theta \prod_{k = t}^{T-1}(\alpha \vec{U}^\top \vec{D}_{k+1}^\theta + \beta \vec{I})$ and similarly $\A_t^{\theta+\delta} := \vec{D}_t^{\theta+\delta}\prod_{k = t}^{T-1}(\alpha (\vec{U}+\delta_{\vec{U}})^\top \vec{D}_{k+1}^{\theta+\delta} + \beta \vec{I})$. Using this, we have,
\begin{small}
\begin{align}
&\left\lVert\frac{\partial L}{\partial \vec{W}} (\theta) - \frac{\partial L}{\partial \vec{W}} (\theta + \delta)\right\rVert_F \nonumber\\
&= \alpha R_{\v x}  \left\Vert\sum_{t=0}^T
\left[
c(\theta)\cdot\A_t^\theta
\vec{v}
-
c(\theta+\delta)\cdot\A_t^{\theta+\delta}
(\vec{v} + \delta_{\vec{v}})
\right]\right\Vert_2\nonumber\\
&\leq \alpha R_{\v x}\left(
|c(\theta) - c(\theta+\delta)|\cdot \left\Vert\sum_{t=0}^T \A_t^\theta \v v\right\Vert_2
+ \left\Vert \sum_{t=0}^T \A_t^\theta \v v - \A_t^{\theta+\delta}
(\vec{v} + \delta_{\vec{v}})\right\Vert_2
\right)\nonumber\\
&\leq \alpha R_{\v x}\left(
|c(\theta) - c(\theta+\delta)|\cdot \left\Vert\sum_{t=0}^T \A_t^\theta \v v\right\Vert_2
+\left\Vert \sum_{t=0}^T (\A_t^\theta - \A_t^{\theta+\delta}
)\vec{v} \right\Vert_2
+\left\Vert \sum_{t=0}^T \A_t^{\theta+\delta}
\delta_\vec{v} \right\Vert_2
\right)\nonumber \\
&\leq \alpha R_{\v x}\left(
|c(\theta) - c(\theta+\delta)|\cdot R_\vec{v} \left\Vert\sum_{t=0}^T \A_t^\theta\right\Vert_2
+R_\vec{v}\left\Vert\sum_{t=0}^T\A_t^\theta - \A_t^{\theta+\delta}
 \right\Vert_2
+\|\delta_\vec{v}\|_2\left\Vert \sum_{t=0}^T \A_t^{\theta+\delta}\right\Vert_2
\right).
\label{eq:dl2}
\end{align}
\end{small}

We will proceed by bounding the first term in the above equation. Consider,
\begin{small}
  \begin{align}\label{eq:at_bound}
    \left\Vert\sum_{t=0}^T \A_t^\theta\right\Vert_2
    &\leq \D \sum_{t=0}^T \left\Vert\prod_{k=t}^{T-1}(\alpha \vec{U}^\top \vec{D}_{k+1}^\theta + \beta \vec{I}) \right\Vert_2\nonumber\\
    &\leq\D \sum_{t=0}^T \left( \alpha\D\cdot \|\v U \|_2 +\beta \right)^{T-t}\nonumber \\
    &\leq \D \frac{|(\alpha\D\cdot \|\v U \|_2 +\beta )^{T+1}-1|}{|\alpha\D\cdot \|\v U \|_2 +\beta -1|}\nonumber \\
    &\stackrel{\zeta_1}{\leq} \D\frac{(1+\alpha\cdot(\D\|U\|_2-1))^{T+1}-1}{\alpha |\D\|\v U\|_2-1|}\nonumber\\
    &\stackrel{\zeta_2}{\leq} 2 \D \cdot(T+1),
  \end{align}
\end{small}
where $\zeta_1$ follows by setting $\beta = 1-\alpha$ and $\zeta_2$ follows by using the inequality $(1+x)^r \leq 1+2rx$ for $(r-1)x\leq 1/2$ and the fact that $\alpha \leq \frac{1}{4T\cdot|\D\|U\|_2 -1| }$. Note that the third term in Equation \eqref{eq:at_bound} can be bounded in a similar way as above by $2 \D \cdot(T+1)$ using $\alpha \leq \frac{1}{4T\cdot R_\vec{U}}$. We now proceed to bound the second term. Consider the following for any fixed value of $t$,
\begin{small}
  \begin{align*}
    \left\Vert\A_t^\theta - \A_t^{\theta+\delta}
     \right\Vert_2 &=\left\Vert \vec{D}_t^\theta \prod_{k = t}^{T-1}(\alpha \vec{U}^\top \vec{D}_{k+1}^\theta + \beta \vec{I}) - \vec{D}_t^{\theta+\delta}\prod_{k = t}^{T-1}(\alpha (\vec{U}+\delta_{\vec{U}})^\top \vec{D}_{k+1}^{\theta+\delta} + \beta \vec{I}) \right\Vert_2\\
     &\leq \left\Vert (\vec{D}_t^\theta - \vec{D}_t^{\theta+\delta}) \prod_{k = t}^{T-1}(\alpha \vec{U}^\top \vec{D}_{k+1}^\theta + \beta \vec{I}) \right\Vert_2 + \D\left\Vert \prod_{k = t}^{T-1}(\alpha \vec{U}^\top \vec{D}_{k+1}^\theta + \beta \vec{I}) - \prod_{k = t}^{T-1}(\alpha (\vec{U}+\delta_{\vec{U}})^\top \vec{D}_{k+1}^{\theta+\delta} + \beta \vec{I}) \right\Vert_2\\
     &\leq \left\Vert \vec{D}_t^\theta - \vec{D}_t^{\theta+\delta}\right\Vert_2\cdot\left\Vert \prod_{k = t}^{T-1}(\alpha \vec{U}^\top \vec{D}_{k+1}^\theta + \beta \vec{I}) \right\Vert_2 + \D\left\Vert \prod_{k = t}^{T-1}(\alpha \vec{U}^\top \vec{D}_{k+1}^\theta + \beta \vec{I}) - \prod_{k = t}^{T-1}(\alpha (\vec{U}+\delta_{\vec{U}})^\top \vec{D}_{k+1}^{\theta+\delta} + \beta \vec{I}) \right\Vert_2\\
     &\leq \left\Vert \vec{D}_t^\theta - \vec{D}_t^{\theta+\delta}\right\Vert_2\cdot(\alpha\| \vec{U}\|_2 \D +\beta)^{T-t}
     + \D \underbrace{\left\Vert \prod_{k = t}^{T-1}(\alpha \vec{U}^\top \vec{D}_{k+1}^\theta + \beta \vec{I}) - \prod_{k = t}^{T-1}(\alpha (\vec{U}+\delta_{\vec{U}})^\top \vec{D}_{k+1}^{\theta+\delta} + \beta \vec{I}) \right\Vert_2}_{(I)}.
  \end{align*}
\end{small}
Let $\Delta^\theta_k := \vec{D}_{k}^\theta - \vec{D}_{k}^{\theta+\delta}$.  We will later show that $\left\Vert \Delta^\theta_k\right\Vert_2 \leq \Delta_\theta$ independent of the value of $k$. We focus on term $(I)$ in the expression above:
\begin{small}
  \begin{align}\label{eq:bnd_term1}
&\left\Vert \prod_{k = t}^{T-1}(\alpha \vec{U}^\top \vec{D}_{k+1}^\theta + \beta \vec{I}) - \prod_{k = t}^{T-1}(\alpha (\vec{U}+\delta_{\vec{U}})^\top \vec{D}_{k+1}^{\theta+\delta} + \beta \vec{I}) \right\Vert_2 \nonumber\\
&\quad \leq \left\Vert \prod_{k = t}^{T-1}(\alpha \vec{U}^\top \vec{D}_{k+1}^{\theta+\delta} + \beta \vec{I} +\alpha \vec{U}^\top \Delta^\theta_{k+1})
- \prod_{k = t}^{T-1}(\alpha \vec{U}^\top \vec{D}_{k+1}^{\theta+\delta} + \beta \vec{I} + \alpha \delta_{\vec{U}}^\top \vec{D}_{k+1}^{\theta+\delta}) \right\Vert_2.
  \end{align}
\end{small}
Let $\B_k : = \alpha \vec{U}^\top \vec{D}_{k}^{\theta+\delta} + \beta \vec{I}$, $\C_k := \alpha \vec{U}^\top \Delta^\theta_{k+1}$ and $\G_k := \alpha \delta_{\vec{U}}^\top \vec{D}_{k+1}^{\theta+\delta}$. Note that we have the following bounds on the operator norms of these matrices:
\begin{equation}
  \| \B_k\|_2 \leq \alpha \D\cdot \|\vec{U}\|_2 + \beta=\B_{\max}, \quad \| \C_k\|_2 \leq \alpha \Delta_\theta\cdot \|\vec{U} \|_2=\C_{\max}, \quad \|\G_k\| \leq \alpha\D \cdot \| \delta_\vec{U}\|_2=\G_{\max}.
  \end{equation}
  By our assumptions on $\alpha$, $\B_k$ is invertible and $I+\B_k \C_k \B_k^{-1}$, $I+\B_k \G_k \B_k^{-1}$ are diagonizable. Moreover, $\|\B_k^{-1}\|\leq  2\alpha \D\cdot \|\vec{U}\|_2 + \beta=\B_{\max}^{-1}$. 
  
  Hence, we can rewrite Equation \eqref{eq:bnd_term1} as,
\begin{small}
  \begin{align}\label{eq:bnd_bcg}
    &\left\Vert \prod_{k = t}^{T-1}(\alpha \vec{U}^\top \vec{D}_{k+1}^\theta + \beta \vec{I}) - \prod_{k = t}^{T-1}(\alpha (\vec{U}+\delta_{\vec{U}})^\top \vec{D}_{k+1}^{\theta+\delta} + \beta \vec{I}) \right\Vert_2\nonumber \\
    &\quad \leq \left\Vert\prod_{k = t}^{T-1}\left(\B_k +\C_k \right) - \prod_{k = t}^{T-1}\left(\B_k + \G_k \right) \right\Vert \nonumber\\
    &\quad \leq 4 \|\B_t\| \cdot \left\Vert\prod_{k = t}^{T-1}\left(I +\B_t^{-1}\C_k\B_{t+1} \right) - \prod_{k = t}^{T-1}\left(I + \B_t^{-1} \G_k \B_{t+1}\right) \right\Vert \nonumber\\
    &\quad \leq 4 \|\B_t\| \cdot \left((1+\B_{\max}\cdot\C_{\max}\cdot\B_{\max}^{-1})^{T-t}-1+ (1+\B_{\max}\cdot \G_{\max}\cdot \B_{\max}^{-1} )^{T-t}-1\right),
    \end{align}
\end{small}
where $B_T:= I$ and the last equation follows from the following fact: $\|\prod_{k=1}^T (I+C_k) - I\|\leq (\max_k \|C_k\|+1)^T-1$. 
    
    %
%
%
Combining the above term with Equation \eqref{eq:dl2}:
\begin{small}
  \begin{align}
\left\Vert\sum_{t=0}^T\A_t^\theta - \A_t^{\theta+\delta} \right\Vert_2
&\leq \sum_{t=0}^T\left\Vert\A_t^\theta - \A_t^{\theta+\delta}
\right\Vert_2  \nonumber\\
&\leq \Delta_\theta\cdot\sum_{t=0}^T (\alpha\D \cdot\| \vec{U}\|_2  +\beta)^{T-t} \nonumber+ \D\cdot\B_{\max}^{-1}\cdot\B_{\max}\nonumber\\&\cdot\sum_{t=0}^{T}  \left((1+\B_{\max}\cdot\C_{\max}\cdot\B_{\max}^{-1})^{T-t}-1+ (1+\B_{\max}\cdot\G_{\max}\cdot\B_{\max}^{-1})^{T-t}-1\right)\nonumber\\
&\leq \Delta_\theta\cdot\sum_{t=0}^T (\alpha\D \cdot\| \vec{U}\|_2  +\beta)^{T-t}  + 2\D\cdot(\B_{\max}^{-1})^{3}\cdot(\B_{\max})^{3}\cdot T^2 \cdot((\C_{\max})^{2}+(\G_{\max})^{2}) \nonumber\\
&\stackrel{\zeta_1}{\leq} 2\Delta_\theta\cdot(T+1) + 2\D\cdot(\B_{\max}^{-1})^3\cdot(\B_{\max})^{3}\cdot T^2 \cdot((\C_{\max})^{2}+(\G_{\max})^{2}),\label{eq:dl4}
  \end{align}
\end{small}
where $\zeta_1$ follows by summing the geometric series and using the fact that $\alpha \leq \frac{1}{4T\cdot|\D\|U\|_2 -1| }$.

Using the definition of $D_k^\theta  = \text{diag}(\sigma'(\vec{W}\vec{x}_k+\vec{U}\vec{h}_{k-1}))$ from Section \ref{sec:method} of the paper, we obtain a bound on $\Delta^\theta_k$.
\begin{small}
  \begin{align}\label{eq:dkbnd}
    \left\Vert \vec{D}_k^\theta -\vec{D}_k^{\theta+\delta} \right\Vert_2 &\leq 2\left(R_\vec{x}\cdot \|\delta_{\vec{W}}\|_2 + \sqrt{\hat{D}}\cdot \|\delta_{\vec{U}}\|_2 + R_{\vec{U}}\cdot \| \vec{h}_{k-1} - \vec{h}_{k-1}'\|_2\right)\nonumber \\
    &\stackrel{\zeta_1}{\leq} 2\left(R_\vec{x}\cdot \|\delta_{\vec{W}}\|_2 + \sqrt{\hat{D}}\cdot \|\delta_{\vec{U}}\|_2 + R_{\vec{U}}\cdot \frac{2\sqrt{\hat{D}}\cdot  \|\delta_{\vec{U}} \|_2 + 2\|\delta_{\vec{W}}\|_2R_{\v x}}{|\|U\|_2-1|}\right),
  \end{align}
\end{small}
where $\zeta_1$ follows from using the bound from Equation \eqref{eq:bnd_h}. Combining bounds obtained in Equations \eqref{eq:dl2}, \eqref{eq:at_bound}, \eqref{eq:dl4} and \eqref{eq:dkbnd}, we obtain that,
\begin{small}
\begin{multline*}
\left\lVert\frac{\partial L}{\partial \vec{W}} (\theta) - \frac{\partial L}{\partial \vec{W}} (\theta + \delta)\right\rVert_F \leq \mathcal{O}(\alpha T)\cdot\| \delta \|_F , \quad \text{for} \\\quad \alpha \leq \min\left( \frac{1}{4T\cdot|\D\|U\|_2 -1| }, \frac{1}{4T\cdot R_\vec{U}}, \frac{1}{2T\cdot \Delta_\theta\|U\|_2}, \frac{1}{T\cdot |\|{\vec{U}}\|_2-1|} \right)
\end{multline*}
\end{small}
where the $\mathcal{O}$ notation hides \emph{polynomial} dependence of the Lipschitz smoothness constant of $L$ on $R_{\vec{W}}, R_{\vec{U}}, R_{\vec{v}}, R_{\vec{x}}, \|\vec{U}\|_2, \|\vec{W}\|_2$ and the ambient dimensions $D, \hat{D}$.

\textbf{Deviation bound for $\frac{\partial L}{\partial \vec{U}}$}: Following similar arguments as we did above for $\frac{\partial L}{\partial \vec{W}}$, we can derive the perturbation bound for the term $\frac{\partial L}{\partial \vec{U}}$ as
\begin{equation}
\left\Vert \frac{\partial L}{\partial \vec{U}} (\theta) - \frac{\partial L}{\partial \vec{U}} (\theta + \delta)\right\Vert _F = \mathcal{O}(\alpha T)\cdot\|\delta \|_F
\end{equation}
where the $\mathcal{O}$ notation is the same as above.

Using our bounds in corollary 2.2 of \cite{ghadimi2013}, we obtain the following convergence theorem.
\begin{reptheorem}{thm:conv}[Convergence Bound] \label{repthm:conv}
	Let $[(\v X_1, y_1), \dots, (\v X_n, y_n)]$ be the given labeled sequential training data. Let $L(\v \theta)=\frac{1}{n}\sum_i L(\v X_i, y_i; \v \theta)$ be the loss function with $\v \theta=(\v W, \v U, \v v)$ be the parameters of \algs\ architecture \eqref{eq:fastrnn} with $\beta = 1-\alpha$ and $\alpha$ such that
  \begin{equation*}
    \alpha \leq \min\left( \frac{1}{4T\cdot|\D\|\vec{U}\|_2 -1| }, \frac{1}{4T\cdot R_\vec{U}}, \frac{1}{2T\cdot \Delta_\theta\|\vec{U}\|_2}, \frac{1}{T\cdot |\|{\vec{U}}\|_2-1|} \right),
  \end{equation*}
  where $\D = \sup_{\theta, k}\| \vec{D}_k^\theta\|_2$. Then, randomized stochastic gradient descent \cite{ghadimi2013}, a minor variation of SGD, when applied to the data for a maximum of $M$ iteration outputs a solution $\widehat{\v \theta}$ such that:
	$$ \mathbb{E}[\| \nabla_{\v \theta}L(\widehat{\v \theta})\|_2^2 \| ]\leq \mathcal{B}_M := \frac{\mathcal{O}(\alpha T)L(\v \theta_0)}{M} + \left( \bar{D} + \frac{4R_{\v W} R_{\v U} R_{\v v}}{\bar{D}}\right)\frac{\mathcal{O}(\alpha T)}{\sqrt{M}},$$
	where $R_{\v X} = \max_{\v X} \|\v X\|_F\ $ for $\v X = \{\v U, \v W, \v v\}$, $L(\v \theta_0)$ is the loss of the initial classifier, and the step-size of the $k$-th SGD iteration is fixed as: $\gamma_k = \min \left\{ \frac{1}{\mathcal{O}(\alpha T)}, \frac{\bar{D}}{T \sqrt{M}}\right\}, k \in [M],\ \ \bar{D}\geq 0.$
\end{reptheorem}


\subsection{Generalization Bound for \algs}
\label{sec:rad}
In this subsection, we compute the Rademacher complexity of the class of real valued scalar gated recurrent neural networks such that $\|\vec{U}\|_F \leq R_{\vec{U}}, \|\vec{W}\|_F\leq R_{\vec{W}}$. Also the input $\vec{x}_t$ at time step $t$ is assumed to be point-wise bounded $\|\vec{x}_t\|_2 \leq R_{\vec{x}}$ The update equation of \salg is given by
$$
\vec{h}_{t} = \alpha\sigma (\vec{W}\vec{x}_t + \vec{U}\vec{h}_{t-1}) + \beta \vec{h}_{t-1}.
$$

For the purpose of this section, we use the shorthand $\vec{h}_t^i$ to denote the hidden vector at time $t$ corresponding to the $i^{th}$ data point $\mathbf{X}^i$. We denote the Rademacher complexity of a $T$ layer \salg by $\mathcal{R}_n(\mathcal{F}_T)$ evaluated using $n$ data points.
\begin{align*}
  n \mathcal{R}_n(\mathcal{F}_T) &= \ex_\epsilon \left[\sup_{\vec{W}, \vec{U}}\left\Vert \sum_{i=1}^n \epsilon_i \vec{h}_T^i\right\Vert  \right]\\
  &= \ex_\epsilon \left[\sup_{\vec{W}, \vec{U}}\left\Vert \sum_{i=1}^n \epsilon_i \left(\alpha\sigma (\vec{W}\vec{x}_{T}^i + \vec{U}\vec{h}_{T-1}^i) + \beta \vec{h}_{T-1}^i \right)\right\Vert \right]\\
  &\stackrel{\zeta_1}{\leq} \ex_\epsilon \left[\sup_{\vec{W}, \vec{U}}\beta \left\Vert \sum_{i=1}^n \epsilon_i  \vec{h}_{T-1}^i \right\Vert \right]
  + \ex_\epsilon \left[\sup_{\vec{W}, \vec{U}} \alpha \left\Vert\sum_{i=1}^n \epsilon_i  (\sigma (\vec{W}\vec{x}_{T}^i + \vec{U}\vec{h}_{T-1}^i)) \right\Vert \right]\\
  &\stackrel{\zeta_2}{\leq} \beta \mathcal{R}_n(\mathcal{F}_{T-1})
  + 2 \ex_\epsilon \left[\sup_{\vec{W}, \vec{U}} \alpha \left\Vert\sum_{i=1}^n \epsilon_i  (\vec{W}\vec{x}_{T}^i + \vec{U}\vec{h}_{T-1}^i) \right\Vert \right]\\
  &\stackrel{\zeta_3}{\leq} \beta \mathcal{R}_n(\mathcal{F}_{T-1})
  + 2\alpha \ex_\epsilon \left[\sup_{W} \left\Vert\sum_{i=1}^n \epsilon_i  \vec{W}\vec{x}_{T}^i \right\Vert \right]
  + 2 \alpha \ex_\epsilon \left[\sup_{\vec{W}, \vec{U}} \left\Vert\sum_{i=1}^n \epsilon_i  \vec{U}\vec{h}_{T-1}^i) \right\Vert \right]\\
  &\stackrel{\zeta_4}{\leq} \beta \mathcal{R}_n(\mathcal{F}_{T-1})
  + 2\alpha R_{\vec{W}}\ex_\epsilon \left[ \left\Vert\sum_{i=1}^n \epsilon_i \vec{x}_{T}^i \right\Vert \right]
  + 2 \alpha R_{\vec{U}}\ex_\epsilon \left[\sup_{\vec{W}, \vec{U}} \left\Vert\sum_{i=1}^n \epsilon_i  \vec{h}_{T-1}^i) \right\Vert \right]\\
  &
  \leq (\beta + 2\alpha R_{\vec{U}}) \mathcal{R}_n(\mathcal{F}_{T-1}) + 2\alpha R_{\vec{W}} R_{\vec{X}}\sqrt{n}\\
  &\leq (\beta + 2\alpha R_{\vec{U}})^2 \mathcal{R}_n(\mathcal{F}_{T-2}) + 2\alpha R_{\vec{W}} R_{\vec{X}}\sqrt{n}(1 + (\beta + 2\alpha R_{\vec{U}}))\\
  &\vdots\\
  &\leq 2\alpha R_{\vec{W}} R_{\vec{X}} \sum_{t=0}^{T-1} \left( \beta + 2\alpha R_{\vec{U}} \right)^{T-t}\sqrt{n}\\\
  &\leq 2\alpha R_{\vec{W}} R_{\vec{X}}\left( \frac{\left( \beta + 2\alpha R_{\vec{U}} \right)^{T+1}-1}{\left( \beta + 2\alpha R_{\vec{U}} \right)-1}\right)\sqrt{n}\\
  &\leq 2\alpha R_{\vec{W}} R_{\vec{X}}\left( \frac{\left( 1 + \alpha( 2R_U-1) \right)^{T+1}-1}{ \alpha(2 R_{\vec{U}} -1)}\right)\sqrt{n} \\
  &\stackrel{\zeta_5}{\leq} 2 R_{\vec{W}} R_{\vec{X}} \left( \frac{2\alpha (2R_{\vec{U}} -1)(T+1)}{(2 R_{\vec{U}} -1)}\right)\sqrt{n},
\end{align*}
where $\zeta_1, \zeta_3$ follows by triangle inequality and noting that the terms in the sum of expectation are pointwise bigger than the previous term, $\zeta_2$ follows from the Ledoux-Talagrand contraction, $\zeta_4$ follows using an argument similar from Lemma $1$ in \cite{golowich2017} and $\zeta_5$ holds for $\alpha \leq \frac{1}{2(2R_{\vec{U}} -1)T}$.

\begin{reptheorem}{thm:gen}[Generalization Error Bound] \label{repthm:gen}
	  \cite{bartlett2002}  Let $\mathcal{Y}, \hat{\mathcal{Y}} \subseteq [0,1]$ and let $\mathcal{F}_T$ denote the class of \algs\ with $\|\v U\|_F \leq R_{\v U}, \|\v W\|_F\leq R_{\v W}$. Let the final classifier be given by $\sigma(\v v^\top \v h_T)$, $\|\v v\|_2 \leq R_{\v v}$ . Let $L:\mathcal{Y}\times  \hat{\mathcal{Y}} \to [0, B]$ be any $1$-Lipschitz loss function. Let $D$ be any distribution on $\mathcal{X}\times \mathcal{Y}$ such that $\|\v x_{it}\|_2 \leq R_x$ a.s. Let $0\leq \delta \leq 1$. For all $\beta = 1-\alpha$ and $alpha$ such that,
    \begin{equation*}
\alpha \leq \min\left( \frac{1}{4T\cdot|\D\|\vec{U}\|_2 -1| }, \frac{1}{4T\cdot R_\vec{U}}, \frac{1}{T\cdot |\|{\vec{U}}\|_2-1|} \right).
\end{equation*}
      where $\D = \sup_{\theta, k}\| \vec{D}_k^\theta\|_2$, we have that with probability at least $1-\delta$, all functions $f \in \v v \circ \mathcal{F}_T$ satisfy,
	  \begin{equation*}
	    \ex_D[L(f(\v X), y)] \leq \frac{1}{{n}}\sum_{i=1}^n L(f(\v X_i), y_i) + \mathcal{C}\frac{O(\alpha T)}{\sqrt{n}} + B\sqrt{\frac{\ln(\frac{1}{\delta})}{n}},
	  \end{equation*}
	  where $\mathcal{C} = R_{\vec{W}} R_{\vec{U}} R_{\vec{x}} R_{\vec{v}}$ represents the boundedness of the parameter matrices and the data.
\end{reptheorem}
The Rademacher complexity bounds for the function class $\mathcal{F}_T$ have been instantiated from the calculations above.

\section{Dataset Information}
\label{sec:datasetinfo}

\textbf{Google-12 \& Google-30:} Google Speech Commands dataset contains 1 second long utterances of 30 short words (30 classes) sampled at 16KHz. Standard log Mel-filter-bank featurization with 32 filters over a window size of 25ms and stride of 10ms gave 99 timesteps of 32 filter responses for a 1-second audio clip. For the 12 class version, 10 classes used in Kaggle's Tensorflow Speech Recognition challenge\footnote{\url{https://www.kaggle.com/c/tensorflow-speech-recognition-challenge}} were used and remaining two classes were noise and background sounds (taken randomly from remaining 20 short word utterances). Both the datasets were zero mean - unit variance normalized during training and prediction.

\textbf{Wakeword-2:} Wakeword-2 consists of 1.63 second long utterances sampled at 16KHz. This dataset was featurized in the same way as the Google Speech Commands dataset and led to 162 timesteps of 32 filter responses. The dataset was zero mean - unit variance normalized during training and prediction.

\textbf{HAR-2}\footnote{\url{https://archive.ics.uci.edu/ml/datasets/human+activity+recognition+using+smartphones}}: Human Activity Recognition (HAR) dataset was collected from an accelerometer and gyroscope on a Samsung Galaxy S3 smartphone. The features available on the repository were directly used for experiments. The 6 activities were merged to get the binarized version. The classes \{Sitting, Laying, Walking\_Upstairs\} and \{Standing, Walking, Walking\_Downstairs\} were merged to obtain the two classes. The dataset was zero mean - unit variance normalized during training and prediction.

\textbf{DSA-19}\footnote{\url{https://archive.ics.uci.edu/ml/datasets/Daily+and+Sports+Activities}}: This dataset is based on Daily and Sports Activity (DSA) detection from a resource-constrained IoT wearable device with 5 Xsens MTx sensors having accelerometers, gyroscopes and magnetometers on the torso and four limbs. The features available on the repository were used for experiments. The dataset was zero mean - unit variance normalized during training and prediction.

\textbf{Yelp-5:}  Sentiment Classification dataset based on the text reviews\footnote{\url{https://www.yelp.com/dataset/challenge}}. The data consists of 500,000 train points and 500,000 test points from the first 1 million reviews. Each review was clipped or padded to be 300 words long. The vocabulary consisted of 20000 words and 128 dimensional word embeddings were jointly trained with the network.

\textbf{Penn Treebank:} 300 length word sequences were used for word level language modeling task using Penn Treebank (PTB) corpus. The vocabulary consisted of 10,000 words and the size of trainable word embeddings was kept the same as the number of hidden units of architecture.

\textbf{Pixel-MNIST-10:} Pixel-by-pixel version of the standard MNIST-10 dataset \footnote{\url{http://yann.lecun.com/exdb/mnist/}}.The dataset was zero mean - unit variance normalized during training and prediction.

\textbf{AmazonCat-13K~\citep{mcauley2013hidden, xmlcode}:} AmazonCat-13K is an extreme multi-label classification dataset with 13,330 labels. The raw text from title and content for Amazon products was provided as an input with each product being assigned to multiple categories. The input text was clipped or padded to ensure that it was 500 words long with a vocabulary of size 267,134. The 50 dimensional trainable word embeddings were initialized with GloVe vectors trained on Wikipedia.

\section*{Evaluation on Multilabel Dataset}
The models were trained on the AmazonCat-13K dataset using Adam optimizer with a learning rate of 0.009 and batch size of 128. Binary Cross Entropy loss was used where the output of each neuron corresponds to the probability of a label being positive. 128 hidden units were chosen across architectures and were trained using PyTorch framework.

\begin{table}[h!]
\centering
\caption{Extreme Multi Label Classification}
\label{tab:XML}
\begin{tabular}{@{}lcccccc@{}}
\toprule
Dataset & \multicolumn{6}{c}{AmazonCat - 13K}                                                                         \\ \midrule
        & P@1   & P@2   & P@3   & P@4   & P@5   & \begin{tabular}[c]{@{}c@{}}Model Size - \\ RNN (KB)\end{tabular} \\ \midrule
GRU     & 92.82 & 85.18 & 77.09 & 69.42 & 61.85 & 268                                                            \\
RNN     & 40.24 & 28.13 & 22.83 & 20.29 & 18.25 & 89.5                                                             \\
\algfloat & 92.66 & 84.67 & 76.19 & 66.67 & 60.63 & 90.0                                                               \\
\salg   & 91.03 & 81.75 & 72.37 & 64.13 & 56.81 & 89.5                                                             \\
UGRNN   & 92.84 & 84.93 & 76.33 & 68.27 & 60.63 & 179                                                              \\ \bottomrule
\end{tabular}
\end{table}

The results in Table~\ref{tab:XML} show that \algfloat achieves classification performance similar to state-of-the-art gated architectures (GRU, LSTM) while still having 2-3x lower memory footprint. Note that the model size reported doesn't include the embeddings and the final linear classifier which are memory intensive when compared to the model itself. \salg, as shown in the earlier experiments, stabilizes standard RNN and achieves an improvement of over 50\% in classification accuracy (P@1).

\section{Supplementary Experiments}\label{sec:supp_exp}

\textbf{Accuracy vs Model Size:} This paper evaluates the trade-off between model size (in the range 0-128Kb) and accuracy across various architectures.

\begin{figure}[h!]
\centering
  \includegraphics[width=0.7\linewidth, height=0.3\linewidth]{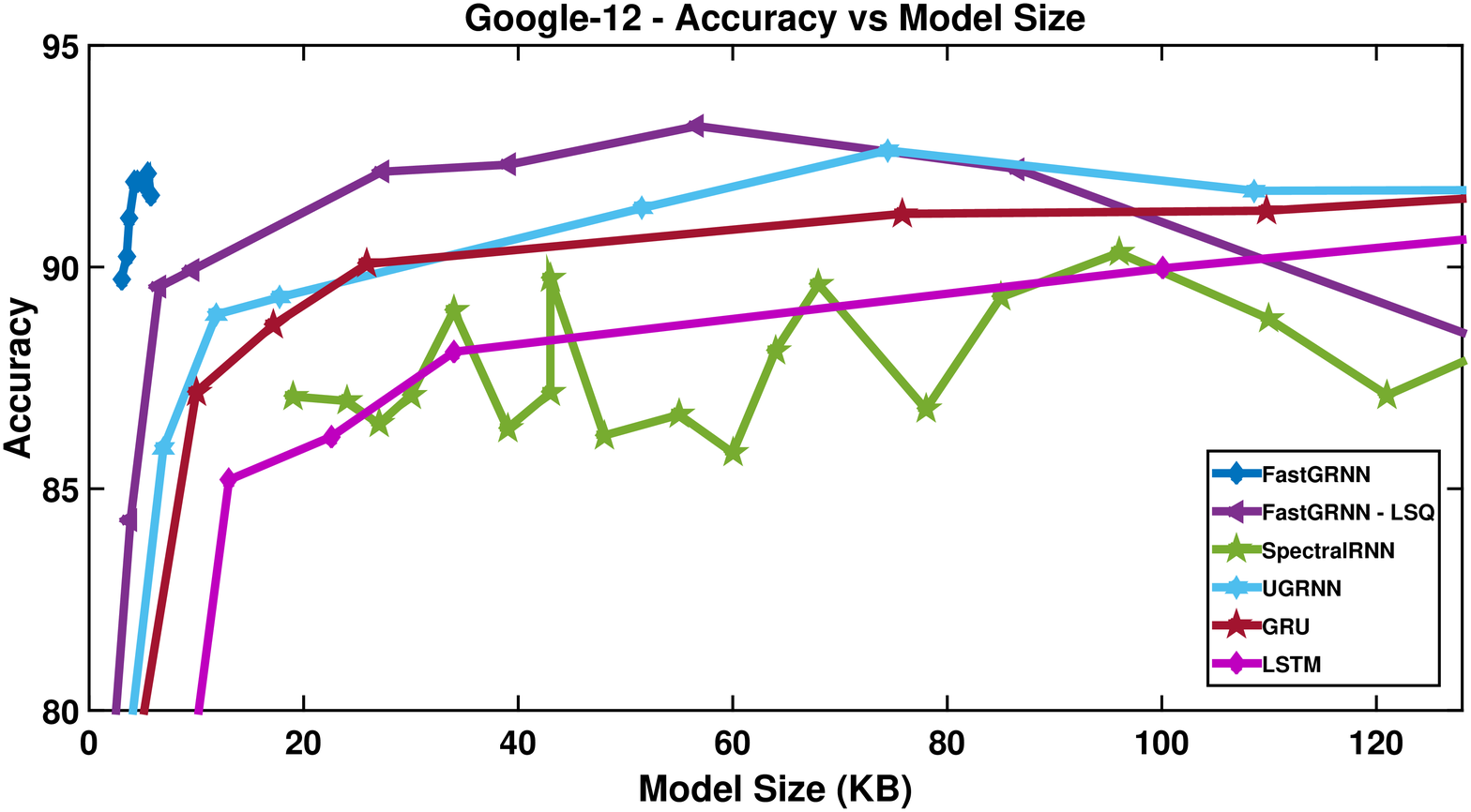}
  \caption{Accuracy vs Model Size}
  \label{fig:G12}
\end{figure}

\begin{figure}[h!]
\centering
  \includegraphics[width=0.7\linewidth, height=0.3\linewidth]{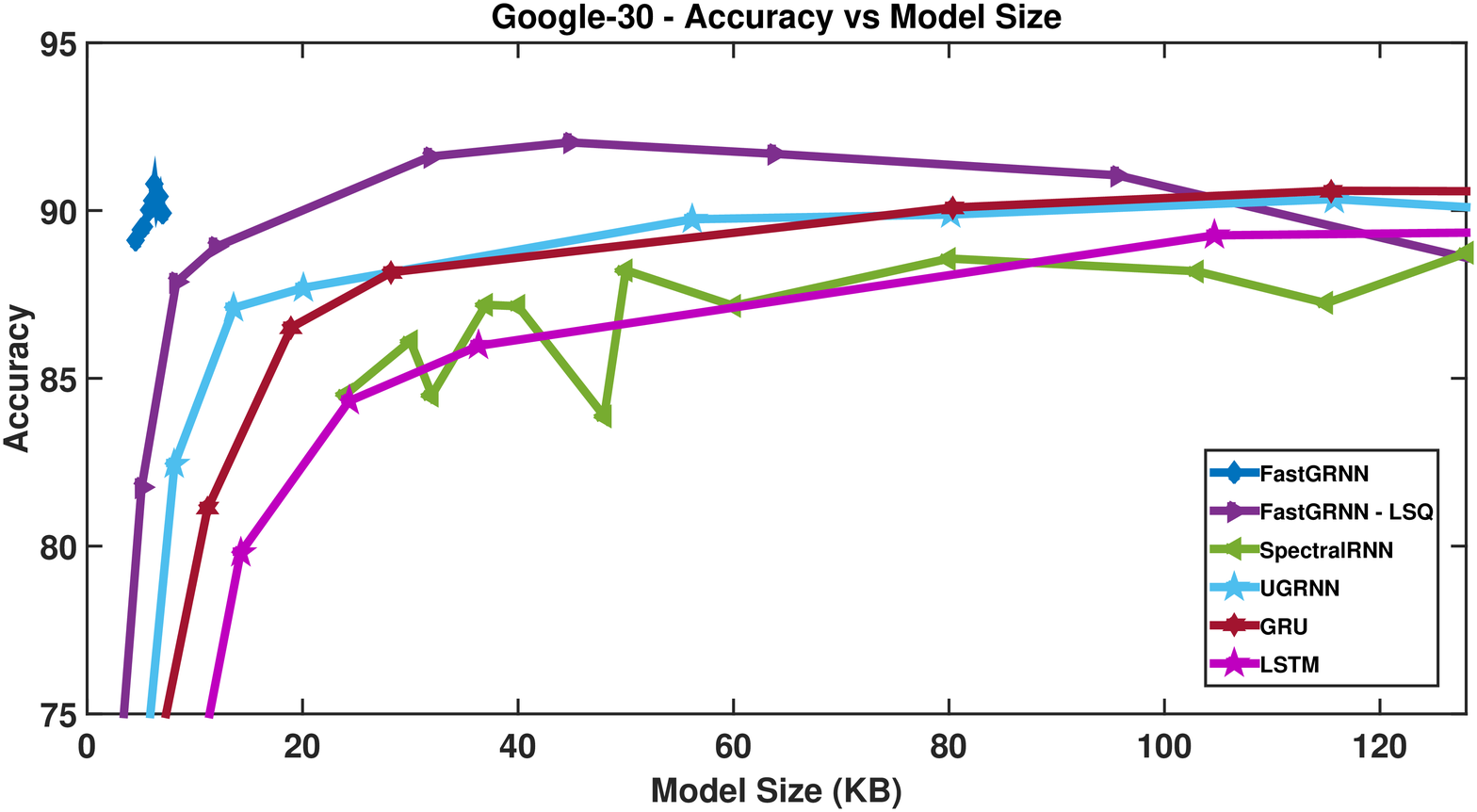}
  \caption{Accuracy vs Model Size}
  \label{fig:G30}
\end{figure}

Figures~\ref{fig:G12},~\ref{fig:G30} show the plots for analyzing the model-size vs accuracy trade-off for \alg, \algfloat along with leading unitary method SpectralRNN and the gated methods like UGRNN, GRU, and LSTM. \alg is able to achieve state-of-the-art accuracies on Google-12 and Google-30 datasets at significantly lower model sizes as compared to other baseline methods.

\begin{figure}[h!]
\centering
  \includegraphics[width=0.7\linewidth, height=0.3\linewidth]{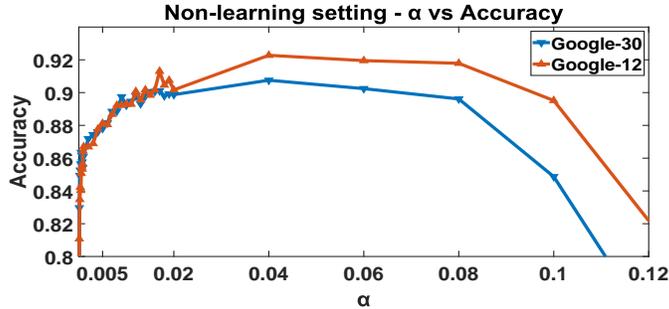}
  \caption{Accuracy vs $\alpha$ in non-learning setting where the parameters of the classifier was learnt and evaluated for a range of fixed $\alpha$ values (using 99 timesteps).}
  \label{fig:alphavsAcc}
\end{figure}

\textbf{Bias due to the initial hidden states:} In order to understand the bias induced at the output by the initial hidden state $h_0$, we evaluated a trained \salg classifier on the Google-12 dataset with 3 different initializations sampled from a standard normal distribution. The resulting accuracies had a mean value of $92.08$ with a standard deviation of $0.09$, indicating that the initial state does not induce a bias in \salg prediction in the learning setting. In the non-learning setting, the initial state can bias the final solution for very small values of $\alpha$. Indeed, setting $\alpha=0$ and $\beta=1$ will bias the final output to the initial state. However, as Figure~\ref{fig:alphavsAcc}indicates, such an effect is observed only for extremely small values of $\alpha \in (0, 0.005)$. In addition, there is a large enough range for $\alpha \in (0.005, 0.08)$ where the final output of \salg is not biased and is easily learnt by \salg.

\textbf{$\alpha$ and $\beta$ of FastRNN:} $\alpha$ and  $\beta$ are the trainable weights of the residual connection in FastRNN. Section~\ref{sec:analysis} shows that FastRNN has provably stable training for the setting of $\alpha/\beta=O(1/T)$. Table~\ref{tab:alphaT} shows the learnt values of $\alpha$ and $\beta$ for various timesteps ($T$) across 3 datasets.
\begin{table}[h!]
\centering
\vspace{-3mm}
\caption{Scaling of $\alpha$ and $\beta$ vs Timesteps for \salg with $\tanh$ non-linearity: With $\alpha$ set as a trainable parameter, it scales as $O(1/T)$ with the number of timesteps as suggested by Theorem~\ref{thm:conv}.}
\label{tab:alphaT}
\begin{tabular}{@{}ccccccccc@{}}
\toprule
\multicolumn{3}{c}{Google-12}                    & \multicolumn{3}{c}{HAR-2}                    & \multicolumn{3}{c}{MNIST-10} \\ \midrule
Timesteps & $\alpha$  & \multicolumn{1}{c}{$\beta$}   & Timesteps & $\alpha$  & \multicolumn{1}{c}{$\beta$}   & Timesteps  & $\alpha$  & $\beta$   \\ \midrule
99        & 0.0654 & \multicolumn{1}{c}{0.9531} & 128       & 0.0643 & \multicolumn{1}{c}{0.9652} & 112        & 0.0617 & 0.9447 \\
33        & 0.2042 & \multicolumn{1}{c}{0.8898} & \phantom{0}64        & 0.1170  & \multicolumn{1}{c}{0.9505} & \phantom{0}56         & 0.1193 & 0.9266 \\
11        & 0.5319 & \multicolumn{1}{c}{0.7885} & \phantom{0}32        & 0.1641 & \multicolumn{1}{c}{0.9606} & \phantom{0}28         & 0.2338 & 0.8746 \\
\phantom{0}9         & 0.5996 & \multicolumn{1}{c}{0.7926} & \phantom{0}16        & 0.2505 & \multicolumn{1}{c}{0.9718} & \phantom{0}14         & 0.3850  & 0.8251 \\
\phantom{0}3         & 0.6878 & \multicolumn{1}{c}{0.8246} & \phantom{00}8         & 0.3618 & \multicolumn{1}{c}{0.9678} & \phantom{00}7          & 0.5587 & 0.8935 \\ \bottomrule
\end{tabular}
\end{table}
\vspace{-3mm}

\section{Compression Components of \alg}
The Compression aspect of \alg has 3 major components: 1) Low-rank parameterization (\textbf{L}) 2) Sparsity (\textbf{S}) and 3) Byte Quantization (\textbf{Q}). The general trend observed across dataset is that low-rank parameterization increase classification accuracies while the sparsity and quantization help reduced the model sizes by 2x and 4x respectively across datasets.

\begin{figure*}[h!]
\centering\hspace*{-3ex}
\vspace{-3mm}
\begin{tabular}{cc}
  \hspace{-7ex}\includegraphics[width=\linewidth, height=0.5\linewidth]{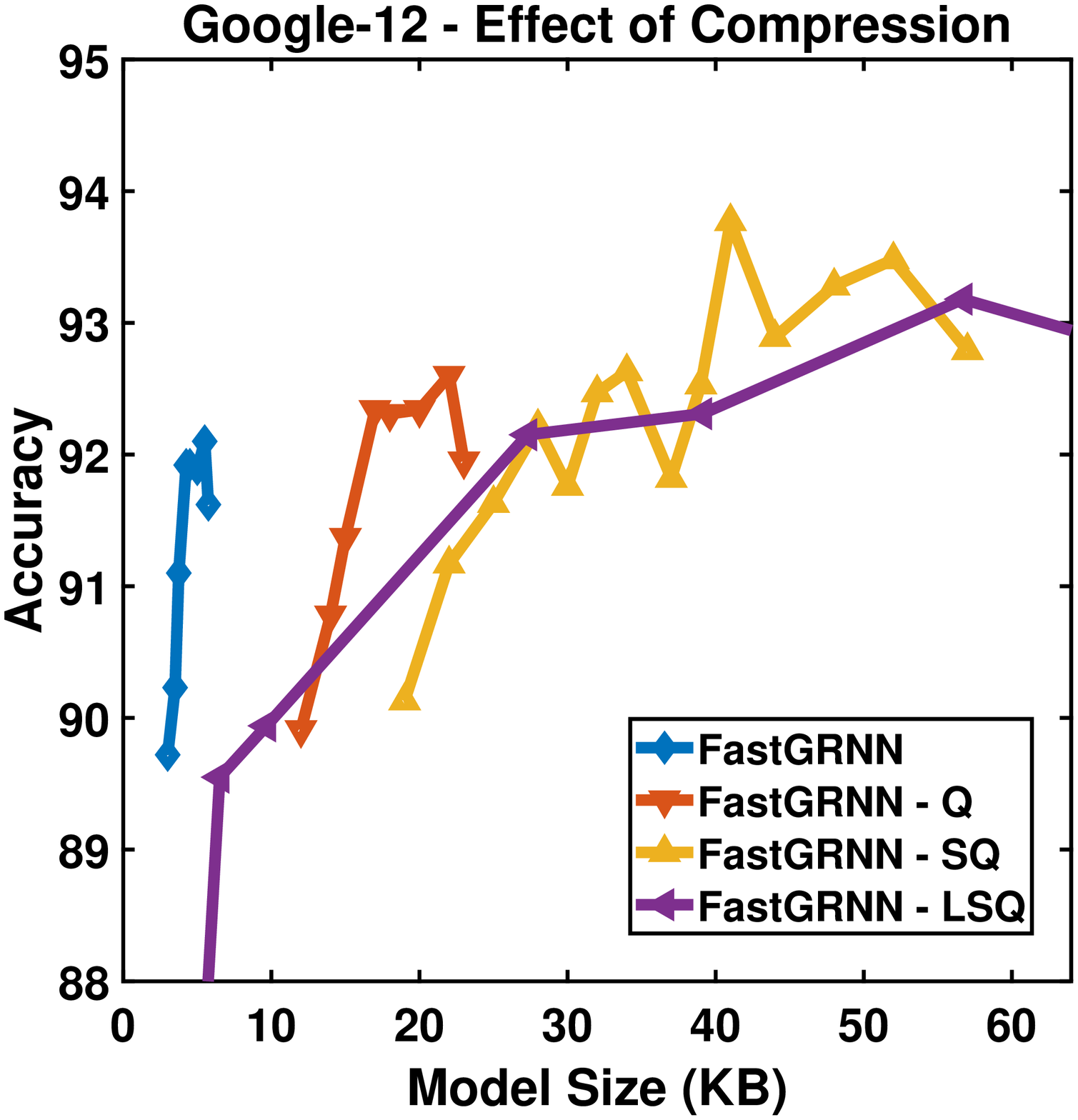}&
    \hspace{-40ex}
  \includegraphics[width=\linewidth, height=0.5\linewidth]{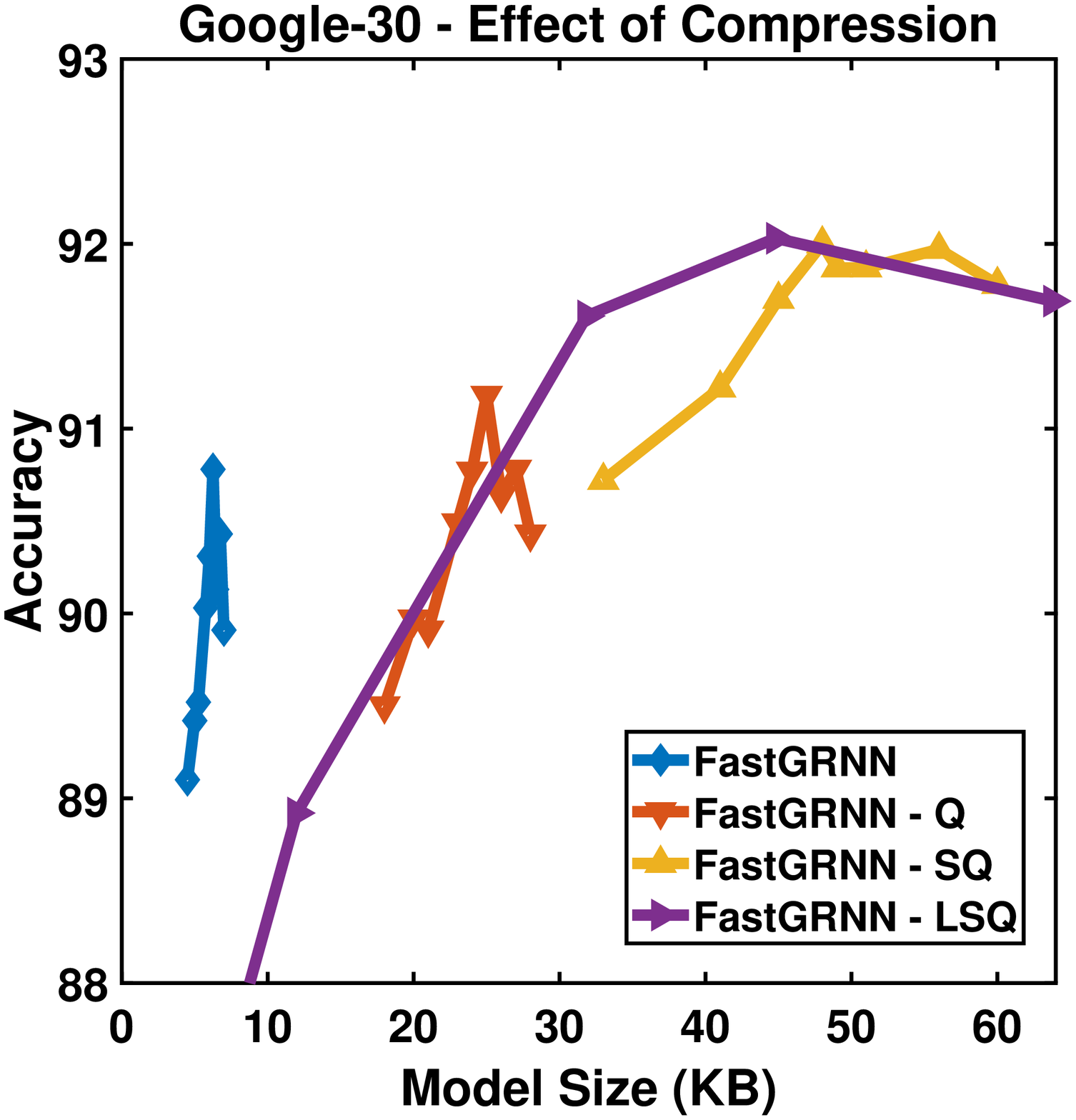}\\
\hspace{-40ex}(a)&\hspace{-70ex}(b)
\end{tabular}
\caption{\small{Figures (a) and (b) show the effect of LSQ components over the model size range of 0-64KB.}}
  \label{fig:plt_com}
\end{figure*}

Tables~\ref{tab:cRes},~\ref{tab:cWake} and~\ref{tab:cPTB} show the trend when each of the component is gradually removed from \alg to get to \algfloat. Note that the hyperparameters have been re-tuned along with the relevant constraints to obtain each model in the table. Figure ~\ref{fig:plt_com} shows the effect of each of LSQ components for two Google datasets.

\begin{table}[h!]
\centering
\makebox[0pt][c]{\parbox{1.1\textwidth}{%
\begin{minipage}[b]{1.0\hsize}\centering
    \captionsetup{font=small}
       \caption{Components of Compression}

  \resizebox{0.98\linewidth}{!}
  {
\begin{tabular}{@{}lcrcrcrcc@{}}
\toprule
Dataset                 & \multicolumn{2}{c}{\alg}                                                                                                         & \multicolumn{2}{c}{\alg-Q}& \multicolumn{2}{c}{\alg-SQ}                                                                                                         & \multicolumn{2}{c}{\algfloat}                                                                                   \\ \midrule
                             & \begin{tabular}[c]{@{}c@{}}Accuracy\\ (\%)\end{tabular} & \multicolumn{1}{c}{\begin{tabular}[c]{@{}c@{}}Model \\ Size (KB)\end{tabular}} & \begin{tabular}[c]{@{}c@{}}Accuracy\\ (\%)\end{tabular} & \multicolumn{1}{c}{\begin{tabular}[c]{@{}c@{}}Model \\ Size (KB)\end{tabular}} & \begin{tabular}[c]{@{}c@{}}Accuracy\\ (\%)\end{tabular} & \multicolumn{1}{c}{\begin{tabular}[c]{@{}c@{}}Model \\ Size (KB)\end{tabular}} & \begin{tabular}[c]{@{}c@{}}Accuracy\\ (\%)\end{tabular} & \multicolumn{1}{c}{\begin{tabular}[c]{@{}c@{}}Model \\ Size (KB)\end{tabular}} \\ \midrule
\multirow{1}{*}{Google-12}   & 92.10                                                   & \multicolumn{1}{c}{5.50} & 92.60                                                   & \multicolumn{1}{c}{22} & 93.76                                                   & \multicolumn{1}{c}{\phantom{0}41}                                                         & 93.18                                                    & \phantom{0}57                                                         \\

\multirow{1}{*}{Google-30}   & 90.78                                                   & \multicolumn{1}{c}{6.25} & 91.18                                                   & \multicolumn{1}{c}{25} & 91.99                                                   & \multicolumn{1}{c}{\phantom{0}38}                                                         & 92.03                                                   & \phantom{0}45                                                         \\

\multirow{1}{*}{HAR-2}   & 95.59                                                   & \multicolumn{1}{c}{3.00} &  96.37                                                  & \multicolumn{1}{c}{17} & 96.81                                                   & \multicolumn{1}{c}{\phantom{0}28}                                                         & 95.38                                                   & \phantom{0}29                                                         \\

\multirow{1}{*}{DSA-19}  & 83.73                                                   & \multicolumn{1}{c}{3.25} & 83.93                                                   & \multicolumn{1}{c}{13} & 85.67                                                   & \multicolumn{1}{c}{\phantom{0}22}                                                         & 85.00                                                   & 208                                                        \\

\multirow{1}{*}{Yelp-5}  & 59.43                                                   & \multicolumn{1}{c}{8.00} & 59.61                                                   & \multicolumn{1}{c}{30} & 60.52                                                   & \multicolumn{1}{c}{130}                                                         & 59.51                                                   & 130                                                         \\

\multirow{1}{*}{Pixel-MNIST-10} & 98.20                                                   & \multicolumn{1}{c}{6.00} & 98.58                                                   & \multicolumn{1}{c}{25} & 98.72                                                   & \multicolumn{1}{c}{\phantom{0}37}                                                       & 98.72                                                   & \phantom{0}71                                                         \\

                                                                                   \bottomrule
\end{tabular}

  \label{tab:cRes}
  }
    \end{minipage}

  \hfill
    \begin{minipage}[b]{1.0\hsize}\centering
    \captionsetup{font=small}
       \caption{Components of Compression for Wakeword-2}

  \resizebox{0.98\linewidth}{!}
  {
  \begin{tabular}{@{}ccccccccc@{}}
  \toprule
  Dataset                 & \multicolumn{2}{c}{\alg}                                                                                                         & \multicolumn{2}{c}{\alg-Q}& \multicolumn{2}{c}{\alg-SQ}                                                                                                         & \multicolumn{2}{c}{\algfloat}                                                                              \\ \midrule
                          & \begin{tabular}[c]{@{}c@{}}F1\\ Score\end{tabular} & \multicolumn{1}{c}{\begin{tabular}[c]{@{}c@{}}Model \\ Size (KB)\end{tabular}} & \begin{tabular}[c]{@{}c@{}}F1\\ Score\end{tabular} & \multicolumn{1}{c}{\begin{tabular}[c]{@{}c@{}}Model \\ Size (KB)\end{tabular}} & \begin{tabular}[c]{@{}c@{}}F1\\ Score\end{tabular} & \multicolumn{1}{c}{\begin{tabular}[c]{@{}c@{}}Model \\ Size (KB)\end{tabular}} & \begin{tabular}[c]{@{}c@{}}F1\\ Score\end{tabular} & \multicolumn{1}{c}{\begin{tabular}[c]{@{}c@{}}Model \\ Size (KB)\end{tabular}} \\ \midrule
  \multirow{1}{*}{Wakeword-2} & 97.83                                              & \multicolumn{1}{c}{1} & 98.07                                              & \multicolumn{1}{c}{4} & 98.27                                              & \multicolumn{1}{c}{8}                                                          & 98.19                                              & 8                                                          \\
   \bottomrule
  \end{tabular}

  \label{tab:cWake}
  }
    \end{minipage}

  \hfill
    \begin{minipage}[b]{1.0\hsize}\centering
    \captionsetup{font=small}
       \caption{Components of Compression for PTB}

  \resizebox{0.98\linewidth}{!}
  {
  \begin{tabular}{@{}ccccccccc@{}}
  \toprule
  Dataset                 & \multicolumn{2}{c}{\alg}                                                                                                         & \multicolumn{2}{c}{\alg-Q}& \multicolumn{2}{c}{\alg-SQ}                                                                                                         & \multicolumn{2}{c}{\algfloat}                                                                                     \\ \midrule
                           & \begin{tabular}[c]{@{}c@{}}Test\\ Perplexity\end{tabular} & \multicolumn{1}{c}{\begin{tabular}[c]{@{}c@{}}Model\\ Size (KB)\end{tabular}} & \begin{tabular}[c]{@{}c@{}}Test\\ Perplexity\end{tabular} & \multicolumn{1}{c}{\begin{tabular}[c]{@{}c@{}}Model\\ Size (KB)\end{tabular}} & \begin{tabular}[c]{@{}c@{}}Test\\ Perplexity\end{tabular} & \multicolumn{1}{c}{\begin{tabular}[c]{@{}c@{}}Model\\ Size (KB)\end{tabular}}&   \begin{tabular}[c]{@{}c@{}}Test\\ Perplexity\end{tabular} & \begin{tabular}[c]{@{}c@{}}Model\\ Size (KB)\end{tabular} \\ \midrule
  \multirow{1}{*}{PTB-10000} & 116.11                                                    & \multicolumn{1}{c}{38.5} & 115.71                                                    & \multicolumn{1}{c}{154}& 115.23                                                    & \multicolumn{1}{c}{384}                                                      & 115.92                                                    & 513                                                      \\
                          \bottomrule
  \end{tabular}

  \label{tab:cPTB}
  }
    \end{minipage}

}}
\end{table}

\section{Hyperparameters of \alg for reproducibility:}
Table~\ref{tab:hyp} lists the hyperparameters which were used to run the experiments with a random-seed of 42 on a P40 GPU card with CUDA 9.0 and CuDNN 7.1. One can use the Piece-wise linear approximations of $\tanh$ or sigmoid if they wish to quantize the weights.

\begin{table}[h!]
\centering
\caption{Hyperparameters for reproducibility - \alg-Q}
\label{tab:hyp}
\begin{tabular}{@{}lccccccc@{}}
\toprule
Dataset        & \begin{tabular}[c]{@{}c@{}}Hidden \\ Units\end{tabular} & $r_w$ & $r_u$ & $s_w$   & $s_u$   & Nonlinearity & Optimizer \\ \midrule
Google-12      & 100                                                     & 16 & 25 & 0.30 & 0.30 & sigmoid      & Momentum  \\
Google-30      & 100                                                     & 16 & 35 & 0.20 & 0.20 & \phantom{000}$\tanh$         & Momentum  \\
Wakeword-2     & \phantom{0}32                                                      & 10 & 15 & 0.20 & 0.30 & \phantom{000}$\tanh$         & Momentum  \\
Yelp-5         & 128                                                     & 16 & 32 & 0.30 & 0.30 & sigmoid      & \phantom{0000}Adam      \\
HAR-2          & \phantom{0}80                                                      & \phantom{0}5  & 40 & 0.20 & 0.30 & \phantom{000}$\tanh$         & Momentum  \\
DSA-19         & \phantom{0}64                                                      & 16 & 20 & 0.15 & 0.05 & sigmoid      & \phantom{0000}Adam      \\
Pixel-MNIST-10 & 128                                                     &\phantom{0} 1  & 30 & 1.00 & 0.30 & sigmoid      & \phantom{0000}Adam      \\
PTB-10000      & 256                                                     & 64 & 64 & 0.30 & 0.30 & sigmoid      & \phantom{0000}Adam      \\ \bottomrule
\end{tabular}
\end{table}

\section{Timing Experiments on more IoT boards}

Table~\ref{tab:rpi} summarizes the timing results on the Raspberry Pi which has a more powerful processor as compared with Arduino Due. Note that the Raspberry Pi has special instructions for floating point arithmetic
and hence quantization doesn't provide any benefit with respect to compute in this case, apart from bringing down the model size considerably.

\begin{table}[h!]
\centering
\caption{Prediction Time on Raspberry Pi 3 (ms)}
\label{tab:rpi}
\begin{tabular}{@{}lccc@{}}
\toprule
Method  &
\multicolumn{1}{c}{\begin{tabular}[c]{@{}c@{}} Google-12 \end{tabular}}                    & \multicolumn{1}{c}{\begin{tabular}[c]{@{}c@{}}HAR-2\end{tabular}}                    & \multicolumn{1}{c}{\begin{tabular}[c]{@{}c@{}}Wakeword-2\end{tabular}} \\ \midrule

\multirow{1}{*}{\alg}    & \phantom{00}7.7 &\phantom{00}1.8 &\phantom{0}2.5 \\
\multirow{1}{*}{RNN}    & \phantom{0}15.7 & \phantom{00}2.9 &\phantom{0}3.6 \\
\multirow{1}{*}{UGRNN}    & \phantom{0}29.7 & \phantom{00}5.6 &\phantom{0}9.5 \\
\multirow{1}{*}{SpectralRNN}    & 123.2 & 391.0 &17.2 \\ \bottomrule
\end{tabular}
\end{table}

\section{Vectorized \salg}
\label{sec:vectorfastrnn}

As a natural extension of \salg , this paper also benchmarked \salg-vector wherein the scalar $\alpha$ in \salg was extended to a vector and $\beta$ was substituted with $\zeta(1-\alpha)+\nu$ with $\zeta$ and $\nu$ are trainable scalars in $[0,1]$. Tables~\ref{tab:frv1},~\ref{tab:frv2} and~\ref{tab:frv3} summarize the results for \salg-vector and a direct comparison shows that the gating enable \alg is more accurate than \salg-vector. \salg-vector used $\tanh$ as the non-linearity in most cases except for a few (indicated by~\textsuperscript{+}) where ReLU gave slightly better results.

\begin{table}[h!]
\centering
\makebox[0pt][c]{\parbox{1.1\textwidth}{%
    \begin{minipage}[b]{0.5\hsize}\centering
    \captionsetup{font=small}
     \caption{\salg\ Vector - 1}

  \resizebox{0.98\linewidth}{!}
  {
       \begin{tabular}{@{}lccc@{}}
\toprule
Dataset        & \begin{tabular}[c]{@{}c@{}}Accuracy\\ (\%)\end{tabular} & \begin{tabular}[c]{@{}c@{}}Model\\ Size (KB)\end{tabular} & \begin{tabular}[c]{@{}c@{}}Train\\ Time (hr)\end{tabular} \\ \midrule
Google-12      & 92.98\textsuperscript{+}                                                   & \phantom{0}57                                                        & \phantom{0}0.71                                                      \\
Google-30      & 91.68\textsuperscript{+}                                                   & \phantom{0}64                                                        & \phantom{0}1.63                                                      \\
HAR-2          & 95.24\textsuperscript{+}                                                   & \phantom{0}19                                                        & \phantom{0}0.06                                                      \\
DSA-19      & 83.24                                                   & 322                                                       & \phantom{0}0.04                                                      \\
Yelp-5         & 57.19                                                   & 130                                                       & \phantom{0}3.73                                                      \\
Pixel-MNIST-10 & 97.27                                                   & \phantom{0}44                                                        & 13.75                                                     \\ \bottomrule
\end{tabular}
  \label{tab:frv1}
  }
    \end{minipage}
    \hfill
    \begin{minipage}[c]{0.5\hsize}\centering
    \captionsetup{font=small}
       \caption{\salg\ Vector - 2}

  \resizebox{0.8\linewidth}{!}
  {
  \begin{tabular}{@{}lccc@{}}
\toprule
Dataset   & \begin{tabular}[c]{@{}c@{}}F1 \\ Score\end{tabular} & \begin{tabular}[c]{@{}c@{}}Model\\ Size (KB)\end{tabular} & \begin{tabular}[c]{@{}c@{}}Train\\ Time (hr)\end{tabular} \\ \midrule
Wakeword-2 & 97.82                                               & 8                                                         & 0.86                                                      \\ \bottomrule
\end{tabular}
  \label{tab:frv2}
  }

 \caption{\salg\ Vector - 3}
\captionsetup{font=small}
\resizebox{0.98\linewidth}{!}
    {
\begin{tabular}{@{}lcccc@{}}
\toprule
Dataset & \begin{tabular}[c]{@{}c@{}}Test \\ Perplexity\end{tabular} & \begin{tabular}[c]{@{}c@{}}Train \\ Perplexity\end{tabular} & \begin{tabular}[c]{@{}c@{}}Model \\ Size (KB)\end{tabular} & \begin{tabular}[c]{@{}c@{}}Train\\ Time (min)\end{tabular} \\ \midrule
PTB-300 & 126.84                                                     & 98.29                                                       & 513                                                        & 11.7                                                       \\ \bottomrule
\end{tabular}
  \label{tab:frv3}
  }

    \end{minipage}

}}
\end{table}

\section{Effects of Regularization for Language Modeling Tasks}
\label{sec:reg}
This section studies the effect of various regularizations for Language Modeling tasks with the PTB dataset. \cite{merity2017regularizing} achieved state-of-the-art performance on the PTB dataset using a variety of different regularizations and this sections combines those techniqeus with \alg and \alg-LSQ. Table \ref{tab:PTBReg} summarizes the train and test perplexity of \alg. The addition of an extra layer leads to a reduction of 10 points on the test perplexity score as compared to a single layer architecture of \alg. Other regularizations like weight decay and weight dropping also lead to gains of upto 8 points in test perplexity as compared to the baseline \alg architecture, exhibiting that such regularization techniques can be combined with the proposed architectures to obtain better dataset specific performance, especially on the language modelling tasks of the PTB dataset.

The experiments carried out in this paper on the PTB dataset use a sequence length of 300 as compared to those used in  \citep{mikolov2012context, zaremba2014recurrent, inan2016tying, melis2017state, merity2017regularizing} which are generally in the range of 35-70. While standard recurrent architectures are known to work with such short sequence lengths, they typically exhibit unstable behavior in the regime where the sequence lengths are longer. These experiments exhibit the stability properties of \alg (with 256 hidden units) in this regime of long sequence lengths with limited compute and memory resources.



\begin{table}[h!]
\centering
\caption{Language Modeling on PTB - Effect of regularization on \alg}
\label{tab:PTBReg}
\begin{tabular}{@{}lccc@{}}
\toprule
Method  &
\multicolumn{1}{c}{\begin{tabular}[c]{@{}c@{}} Hidden Units \end{tabular}}                    & \multicolumn{1}{c}{\begin{tabular}[c]{@{}c@{}}Test Perplexity\end{tabular}}                    & \multicolumn{1}{c}{\begin{tabular}[c]{@{}c@{}}Train Perplexity\end{tabular}} \\ \midrule

\multirow{1}{*}{1-layer}    & 256 &116.11 & 81.31 \\
\multirow{1}{*}{2-layer}    & 256 & 106.23 & 69.37 \\
\multirow{1}{*}{1-layer + Weight decay}    & 256 & 111.57 & 76.89 \\
\multirow{1}{*}{1-layer + Weight-dropping}    & 256 & 108.56 & 72.46\\
\multirow{1}{*}{1-layer + AR/TAR}    & 256 & 112.78 &78.79 \\ \bottomrule
\end{tabular}
\end{table}


\end{document}